\documentclass{article} % For LaTeX2e
\usepackage{iclr2027_conference,times}
\usepackage{booktabs}    % 用于 \toprule, \midrule, \bottomrule 等优雅的表格线
\usepackage{multirow}    % 用于合并单元格（如 \multirow）
\usepackage{makecell}    % 用于在单元格中换行（如 \makecell）
\usepackage{graphicx}    % 可选，用于调整表格间距等
\usepackage{subcaption}  % 用于并排子图
\usepackage{algorithm}
\usepackage{algpseudocode}
\usepackage{booktabs}
\usepackage{multirow}
\usepackage{makecell}
\usepackage{graphicx}
\usepackage[table]{xcolor}
\usepackage{amssymb} 
\usepackage{wrapfig} % for wrapfigure
\usepackage{graphicx}
\usepackage{float}
\definecolor{bestgreen}{RGB}{220,242,220}
\definecolor{secondyellow}{RGB}{255,244,204}
\usepackage{xcolor}   % 提供 \textcolor
\usepackage{hyperref}
\usepackage{titletoc} % 提供 \startcontents 和 \printcontents

\newcommand{\best}[1]{\cellcolor{bestgreen}\textbf{#1}}
\newcommand{\secondbest}[1]{\cellcolor{secondyellow}\underline{#1}}
\usepackage{amsmath,amsfonts,bm}

\def\eqref#1{equation~\ref{#1}}
\def\1{\bm{1}}

\DeclareMathAlphabet{\mathsfit}{\encodingdefault}{\sfdefault}{m}{sl}
\SetMathAlphabet{\mathsfit}{bold}{\encodingdefault}{\sfdefault}{bx}{n}

\usepackage{url}
\title{PhysForce: Physics-aware Force Control for Interactive Video World Model}
\author{Xuancheng Xu\textsuperscript{1,$\ddagger$}, Xiaofeng Wang\textsuperscript{3}, Yue Ma\textsuperscript{3}, Zheng Zhu\textsuperscript{2,$\dagger$}, Bing-Kun Bao\textsuperscript{4,$\dagger$} \\
\textsuperscript{1}Nanjing University of Posts and Telecommunications \quad 
\textsuperscript{2}GigaAI \\
\textsuperscript{3}Tsinghua University \quad
\textsuperscript{4}Hefei University of Technology \\
Project page: \url{https://xuxuancheng0208.github.io/PhysForce_page/}
}

\iclrfinalcopy % Uncomment for camera-ready version, but NOT for submission.
\begin{document}

\maketitle
\begingroup
\renewcommand{\thefootnote}{\fnsymbol{footnote}}
\footnotetext[2]{Corresponding Author}
\footnotetext[3]{Work done during an internship at GigaAI}
\endgroup
\begin{figure}[h]
  \vspace{-1cm}
  \centering
  \includegraphics[width=0.9\textwidth]{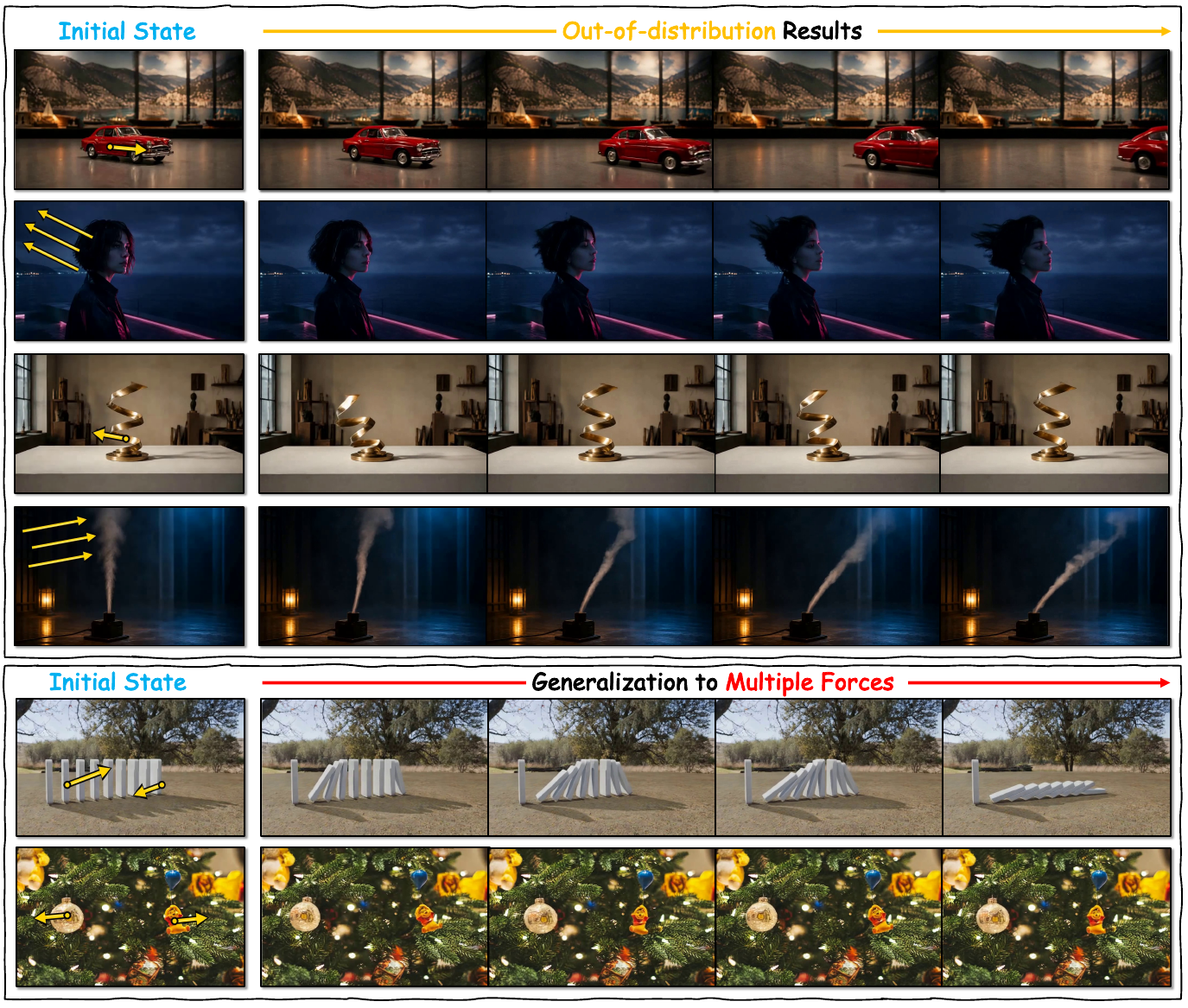}
  \vspace{-5pt}
  \caption{\textbf{Showcase of PhysForce under point and wind forces.} PhysForce generates physically consistent videos across diverse scenes and shows the potential to extend from single force to multiple force control. A single arrow denotes a point force, while three arrows denote wind. Arrow length and orientation indicate the force magnitude and direction respectively.}
  \label{fig:teaser}
\end{figure}

\begin{abstract}
Interactive video world models aim to predict how visual environments evolve in response to user interventions, including physically consistent state changes induced by applied forces.
Existing methods often represent forces as motion trajectories, which may introduce inappropriate motion priors, causing the model to over-rely on encoded displacement patterns.
More fundamentally, the model's internal representations do not reliably localize force-induced state changes, making it difficult to associate applied forces with their resulting physical responses.
Also, limited force-annotated data make it difficult to produce physically consistent and temporally stable predictions in unseen scenarios.
To address these limitations, we propose PhysForce, a two-stage framework that integrates supervised fine-tuning and reinforcement learning through three complementary components.
During supervised fine-tuning, (1) the Latent Force Intervention Encoder maps force location, direction, and magnitude into the video latent space without exposing future motion;
and (2) the Physics Attribution Network associates force interventions with localized changes in the latent state, strengthening the learning of physical responses to applied forces.
During reinforcement learning, (3) Physics-Aware Reinforcement Learning uses self-supervised rewards on out-of-distribution videos to improve physical consistency and generalization.
Extensive experiments show that PhysForce improves force responsiveness, physical consistency, and generalization, demonstrating its effectiveness for interactive video world models.
\end{abstract}

% \begin{wrapfigure}{r}{0.45\textwidth}
% 	\centering
%     \vspace{-10pt}
% 	\includegraphics[width=\linewidth]{figures/introduction.pdf}
%     \vspace{-20pt}
% 	\caption{\footnotesize
% (a) Existing methods encode forces as moving Gaussian blobs, causing trajectory leakage and physically implausible motion.(b) PAN localizes regions affected by the applied force.(c) PARL further improves the physical consistency of generated videos.
% }
%     \label{fig_introduction}
%     \vspace{-0.7cm}
% \end{wrapfigure}
\begin{figure*}[!t]
    \centering
    \includegraphics[width=0.85\textwidth]{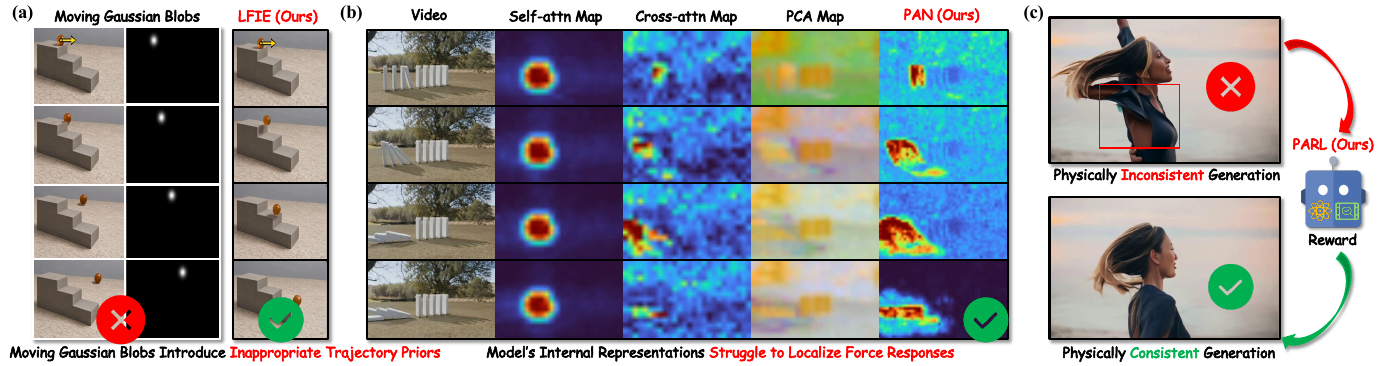}
    \vspace{-10pt}
    \caption{(a) Existing methods encode forces as moving Gaussian blobs, introducing inappropriate trajectory priors that may produce physically implausible responses. (b) Existing methods' internal representations such as self-attention, cross-attention, and PCA features fail to reliably localize force-induced state changes, whereas PAN provides more accurate response localization. (c) PARL further improves the physical consistency of generated videos.}
    \label{fig_introduction}
    \vspace{-10pt}
\end{figure*}
\section{Introduction}
Interactive video world models aim to simulate how visual environments evolve in response to user interventions~\citep{bruce2024genie,chen2026learning,huang2025vid2world}. 
For force-conditioned interaction, the model must predict future states from an initial state and a force defined by its application point, direction, and magnitude, while ensuring that the resulting dynamics are physically plausible and consistent with the applied force.

Existing methods~\citep{gillman2026force,gillman2026goal,wang2026physctrl} introduce force conditioning to model physical interventions, yet face three limitations. 
First, trajectory-based force representations may introduce inappropriate motion priors, causing models to over-rely on encoded displacement patterns rather than infer force induced results, as shown in Fig.~\ref{fig_introduction}(a). 
%First, trajectory-based force representations expose expected displacement, reducing force control to trajectory following rather than learning force induced results, as shown in Fig.~\ref{fig_introduction}(a). 
%Second, standard video-generation objectives capture statistical patterns without explicitly linking force interventions to their effects, unlike humans who acquire intuitive physics through action and observation. 
Second, internal representations such as attention maps and PCA features do not reliably localize force-induced state changes, as shown in Fig.~\ref{fig_introduction}(b). This limits their ability to provide explicit supervision for physical response regions.
Third, limited annotated force data make it difficult to produce physically consistent and temporally stable predictions in unseen scenarios, as shown in Fig.~\ref{fig_introduction}(c).

To address these limitations, we propose PhysForce, a two-stage framework that integrates supervised fine-tuning and reinforcement learning through three complementary components. 
During supervised fine-tuning (SFT), we introduce \textbf{Latent Force Intervention Encoder (LFIE)} to eliminate future-motion leakage from moving Gaussian blobs. In Fig.~\ref{fig_introduction}(a), the leaked trajectory drives the ball along a straight path, overriding its physically plausible motion down the stairs.
LFIE directly maps force location, direction, and magnitude into a learnable latent condition without exposing temporal trajectories or future displacement, requiring the model to infer the resulting state changes.

Second, inspired by how children acquire intuitive physics through interaction and observation~\citep{piaget1952origins,gopnik2004theory}, we introduce the \textbf{Physics Attribution Network (PAN)} to connect force interventions with observed state changes. Self-attention and cross-attention capture feature relationships but fail to reliably trace physical responses, while PCA remains entangled with appearance. PAN instead aligns force-conditioned latent responses with temporal differences to localize where and how strongly state changes occur, as shown in Fig.~\ref{fig_introduction}(b). This supervision makes the relation between intervention and response explicit, encouraging the model to learn how forces drive scene evolution rather than reproduce generic motion patterns.

During reinforcement learning (RL), we introduce \textbf{Physics-Aware Reinforcement Learning (PARL)} to improve physical consistency and generalization to unseen scenes, as shown in Fig.~\ref{fig_introduction}(c). PARL derives an attribution-guided physics reward from the learned dynamics prior of a self-supervised V-JEPA encoder and predictor~\citep{murlabadia2026vjepa2_1,assran2025vjepa2}, reinforcing physically coherent state evolution within PAN-localized response regions. Since the reward can be hacked by generating static videos or visual artifacts, we introduce a Dynamic Gate that evaluates motion, temporal stability, and visual quality to filter such degenerate rollouts.

Extensive evaluations on in-distribution (ID) and out-of-distribution (OOD) benchmarks show that PhysForce improves force controllability, physical consistency, and generalization while preserving visual quality and temporal coherence. 
These results establish PhysForce as an effective framework for physics-aware interaction and state prediction in video world models.

Overall, our contributions are summarized as follows:
\begin{itemize}
    \item We introduce the \textbf{Latent Force Intervention Encoder}, which encodes force location, direction, and magnitude in the video latent space without trajectory leakage, allowing the model to predict state changes caused by the applied force.

    \item We propose the \textbf{Physics Attribution Network}, which learns a force conditioned response map that localizes state changes associated with the intervention.

    \item We develop \textbf{Physics-Aware Reinforcement Learning} to improve physical consistency and generalization using attribution guided self-supervised rewards from videos, with a Dynamic Gate mitigating reward hacking.

    \item PhysForce achieves state-of-the-art force controllability, physical consistency, and generalization while preserving video quality and temporal coherence.
\end{itemize}
\section{Related Works}

\textbf{Video Diffusion Transformer.} Diffusion models~\citep{wang2025lavie,xu2025chain,xu2026processmaker,xu2026fooddiff,mi2026steerface,mi2025morphface} have achieved remarkable success in generating high-quality videos~\citep{blattmann2023stable, singermake, khachatryan2023text2video,ma2024latte}. Leveraging the scalability of the Diffusion Transformer architecture~\citep{peebles2023scalable}, recent models~\citep{yang2024cogvideox,wan2025wan,xu2026breaking} have substantially improved both visual fidelity and motion quality. Building upon these advances, controllable video generation~\citep{ma2025controllable,ma2024followpose,ma2025followcreation,ma2026fastvmt,ma2025followyourmotion,xu2026smrabooth,xu2026discolora,xu2025clgc,wang2026subjectanchor,li2025create,lireactid,li2026multibinder,liu2025javisdit,li2026flashmotion,li2025magicmotion,wang2025cinemaster,wang2026multishotmaster,zhang2025motioncrafter,ma2025consistent,ma2026consistent,zhang2025flexiact} enables generated videos to adhere to specific control conditions, thereby laying a solid foundation for action-conditioned video world models.

\textbf{Physics-aware Video Generation.} Despite the impressive progress of existing video generation models in producing cinematic and visually appealing videos, their ability to faithfully follow real-world physical consistency remains limited. Physics-aware video generation aims to enhance the capability of foundation video generation models to perceive, represent, and generate videos that conform to physical laws. Existing methods can be broadly categorized into several lines. Some approaches~\citep{zhang2026physrvg,gillman2026force,gillman2026goal,liu2024physgen,zhang2024physdreamer,chen2026physstream} introduce implicit physical modeling by converting physical correctness into computable losses or rewards. Other works~\citep{zhang2026videorepa,wang2026prophy,yuan2026inference} align physical representations with model features through external networks or physics priors. Another line of research~\citep{wang2026physctrl,liu2026realwonder,foo2026physical} reconstructs particle fields and combines them with physics solvers, using predicted future particle trajectories to guide video generation. In addition, ~\citep{yuan2025newtongen} explicitly models dynamics and supervises model learning with Newtonian laws of motion.

\textbf{Interactive World Model.} An interactive world model aims to predict future state transitions based on user-specified initial conditions. Recent works~\citep{sun2025worldplay,helios,ye2025yan,he2025matrix,xiang2025pan,shin2026motionstream,gu2026worldcraft,li2025magicworld,wang2026liveedit,wang2026liveact,ma2026livelight,team2026gigaworld} further incorporate control signals, such as camera parameters and trajectories, enabling video models to interactively explore and predict real-world dynamics. However, these control signals typically contain little explicit physical information. In contrast, another line of work~\citep{gillman2026force,gillman2026goal,wang2026streaming,narayanan2026phyco,li2026learning,zhongsurggenesis} introduces physical quantities, such as forces, as initial triggering conditions, with the goal of improving models' understanding of physical interactions and eliciting physically consistent responses to such conditions.

\begin{figure*}[!t]
    \centering
    \includegraphics[width=0.95\textwidth]{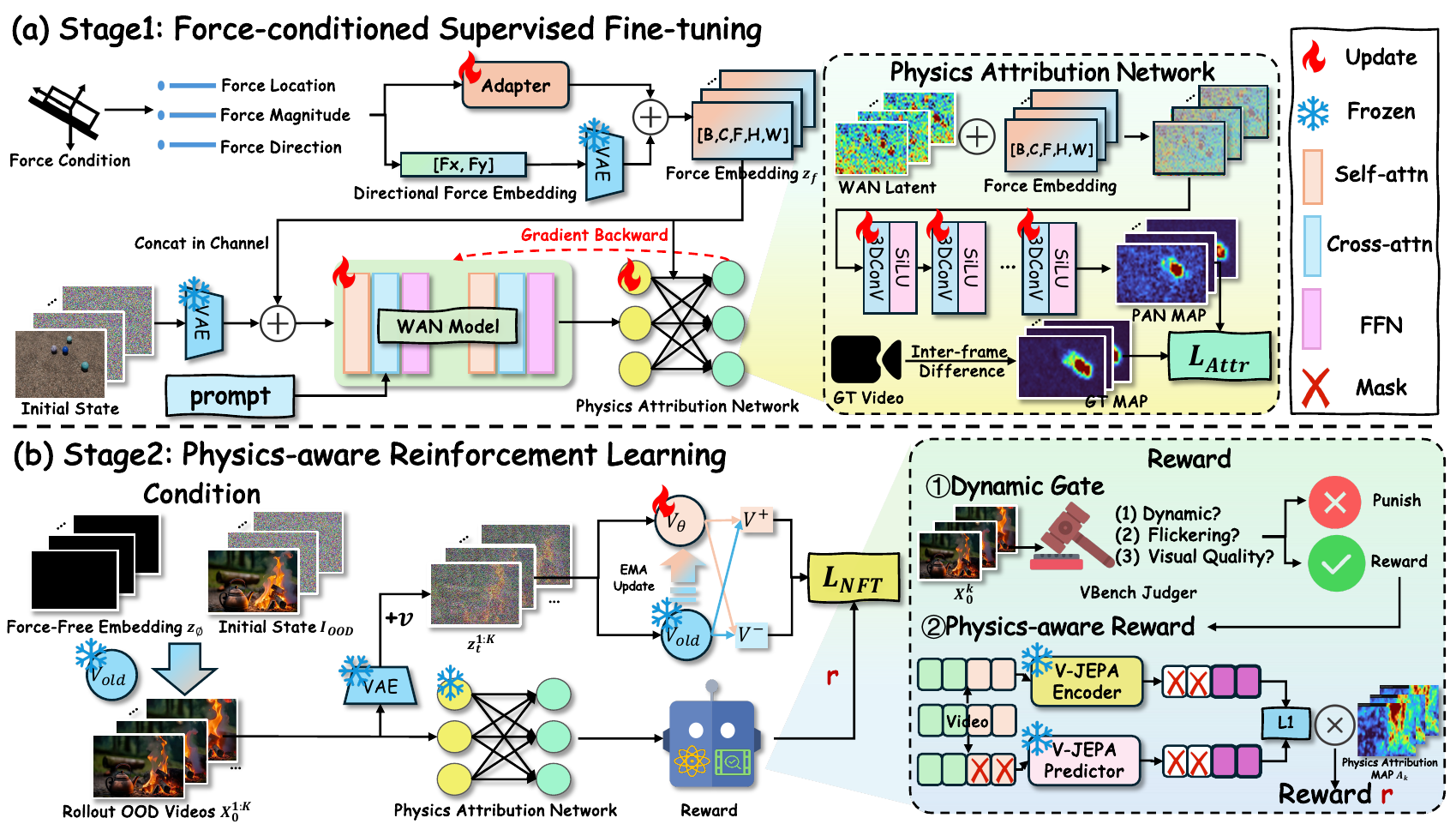}
    \vspace{-10pt}
    \caption{\textbf{Overview of PhysForce.} Stage 1 performs force-conditioned supervised fine-tuning with LFIE and PAN, while Stage 2 applies PARL with a self-supervised reward and Dynamic Gate.}
    \label{fig_pipeline}
    \vspace{-0.5cm}
\end{figure*}
\section{Method}
In this section, we present PhysForce, a two-stage framework for physics-aware force control in interactive video world models. 
The SFT stage learns latent force conditioning and intervention-response attribution, while the reinforcement learning stage improves physical consistency in unseen scenarios. The overall pipeline is shown in Fig.~\ref{fig_pipeline}.
Sec.~\ref{sub:Latent} presents the \textbf{Latent Force Intervention Encoder (LFIE)}, which maps outcome-agnostic force conditions into the video latent space. Sec.~\ref{sub:attritution} introduces the \textbf{Physics Attribution Network (PAN)}, which associates force interventions with localized latent-state changes. Finally, Sec.~\ref{sub:RL} describes \textbf{Physics-aware Reinforcement Learning (PARL)}, which combines attribution-guided self-supervised rewards with a Dynamic Gate to improve unseen-state prediction and mitigate reward hacking.
\noindent\subsection{Preliminary}\label{sub:Preliminary}

\noindent\textbf{Online Diffusion Reinforcement Learning.}
Reinforcement learning approaches~\citep{liu2026flow,guan2026latentrewardregistersdiffusion,cheng2026omniopsdrationaleprivilegedonpolicyselfdistillation,chen2026reflectr1evidencedrivenreflectionselfcorrection,li2026diffusionopd} leverage policy-gradient methods to optimize sequential decision-making processes.
DiffusionNFT~\citep{zheng2025diffusionnft} performs reinforcement learning directly on the forward process by contrasting generated samples based on rewards.

Given a generated sample $x_0$, its condition $c$, and a normalized reward $r(X_0,c)\in[0,1]$, DiffusionNFT constructs implicit positive and negative velocity predictors:
\begin{equation}
\label{eq:implicit-policies}
\begin{aligned}
    v_{\theta}^{+}(z_t,c,t)
    &=(1-\beta)v_{\mathrm{old}}(z_t,c,t)
      +\beta v_{\theta}(z_t,c,t), \\
    v_{\theta}^{-}(z_t,c,t)
    &=(1+\beta)v_{\mathrm{old}}(z_t,c,t)
      -\beta v_{\theta}(z_t,c,t),
\end{aligned}
\end{equation}
where $v_{\mathrm{old}}$ denotes the data-collection policy and $\beta$ controls the reinforcement strength. The model is then optimized using the reward-weighted flow matching objective
\begin{equation}
\label{eq:diffusion-nft-loss}
    \mathcal{L}_{\mathrm{NFT}}
    =
    \mathbb{E}_{X_0,c,t}
    \left[
        r(X_0,c)\left\|v_{\theta}^{+}-v_t\right\|_2^2
        +
        \bigl(1-r(X_0,c)\bigr)
        \left\|v_{\theta}^{-}-v_t\right\|_2^2
    \right].
\end{equation}
where the arguments $(z_t,c,t)$ of $v_\theta^{+}$ and $v_\theta^{-}$ are omitted for brevity. This objective promotes high-reward samples while leveraging low-reward samples as negative feedback. 
%By optimizing directly through the forward flow matching process, DiffusionNFT avoids likelihood estimation and denoising-trajectory storage while remaining compatible with arbitrary sampling solvers.

\noindent\subsection{Stage1: Force-conditioned supervised fine-tuning}
\noindent\subsubsection{Latent Force Intervention Encoder}
\label{sub:Latent}

Trajectory-based force representations specify a sequence of positions and therefore expose the expected motion. To avoid this trajectory leakage, we introduce the LFIE, which represents only the applied force without prescribing its outcome. Given an application point $\mathbf{p}$, a normalized magnitude $s$, and direction $(\cos\theta,\sin\theta)$ in the image coordinate, we construct a four-channel force field:
\begin{equation}
\mathbf{F}(\mathbf{p})=
\left[
M(\mathbf{p}),
W(\mathbf{p})\cos\theta,
W(\mathbf{p})\sin\theta,
W(\mathbf{p})s
\right],
\label{eq:force_map}
\end{equation}
where $M$ is a truncated Gaussian centered at $\mathbf{p}_0$, and $W$ denotes its binary support. The support has a fixed spatial extent, ensuring that force magnitude is represented by $s$ rather than by the size of the force region. For global forces such as wind, we set $M(\mathbf{p})=W(\mathbf{p})=1$. The same force field is repeated along the temporal dimension while its center remains fixed. Consequently, the condition contains no future position, displacement, or object trajectory.

LFIE projects the force condition into the video latent space through two parallel branches. The first branch constructs spatially localized directional components:
\begin{equation}
F_x(\mathbf{p})=M(\mathbf{p})\cdot s \cdot \cos\theta,
\qquad
F_y(\mathbf{p})=M(\mathbf{p})\cdot s \cdot \sin\theta,
\label{eq:directional_components}
\end{equation}
A fixed mapping $\phi$ converts these components into the native control format of the pretrained model. The frozen video VAE then produces the Directional Force Embedding:
\begin{equation}
\mathbf{z}_{\mathrm{DFE}}
=
\mathcal{E}_{VAE}
\left(
\phi(F_x,F_y)
\right),
\label{eq:directional_embedding}
\end{equation}
This branch contains no trainable parameters and preserves compatibility with the pretrained control pathway. In parallel, a trainable adapter $\mathcal{A}_{\psi}$ maps the complete factorized force field directly onto the video latent grid. The two branches are combined by elementwise addition:
\begin{equation}
\mathbf{z}_{f}
=
{\mathbf{z}_{\mathrm{DFE}}}
+
{\mathcal{A}_{\psi}(\mathbf{F})}.
\label{eq:latent_force}
\end{equation}
The output projection of $\mathcal{A}_{\psi}$ is initialized to zero, so LFIE initially follows the frozen directional embedding and progressively learns complementary representations of force location, direction, and magnitude. Finally, the resulting force embedding is concatenated with the visual conditioning latent along the channel dimension.
Since neither branch receives the future motion, the model must infer the resulting state evolution from the interaction between the applied force and the current scene.

\subsubsection{Physics Attribution Network}
\label{sub:attritution}

Standard video-generation objectives supervise the generated video globally without explicitly associating force interventions with their induced state changes. Inspired by Piaget's sensorimotor theory, which relates physical knowledge to actions and their observed consequences~\citep{piaget1952origins}, we introduce the PAN to strengthen intervention-response learning.

At timestep $t$, let $\hat{\mathbf{z}}_t$ and $\mathbf{z}_t^\ast$ denote the predicted and target latent sequences under the same noise realization, and the force embedding $\mathbf{z}_f$ from Sec.~\ref{sub:Latent}. Their temporal responses are defined as:
\begin{equation}
\Delta\hat{\mathbf{z}}_{t,i}
=
\hat{\mathbf{z}}_{t,i+1}
-
\hat{\mathbf{z}}_{t,i},
\qquad
\Delta\mathbf{z}_{t,i}^{\ast}
=
\mathbf{z}_{t,i+1}^{\ast}
-
\mathbf{z}_{t,i}^{\ast}.
\label{eq:temporal_responses}
\end{equation}
Here, $i$ indexes temporal positions in the latent sequence. 
Since the simulated videos are captured by static cameras and their state changes are driven by the applied force, these temporal differences provide dense cues for localizing state changes caused by the applied force.
PAN encodes the predicted response and detached force embedding through separate branches. The force feature modulates the response through a zero-initialized residual gate to produce the attribution feature $\hat{\mathbf{a}}_i$. PAN aligns it with the observed temporal response:
\begin{equation}
\hat{\mathbf{a}}_{i}
=
\operatorname{PAN}
\left(
\Delta\hat{\mathbf{z}}_{i,t},
\operatorname{sg}(\mathbf{z}_{f})
\right),
\qquad
\mathcal{L}_{\mathrm{attr}}
=
\frac{1}{T-1}
\sum_{i=1}^{T-1}
\left\|
\hat{\mathbf{a}}_{i}
-
\Delta\mathbf{z}^{*}_{i,t}
\right\|_{1},
\label{eq:physics_attribution}
\end{equation}
Here, $\operatorname{sg}(\cdot)$ denotes gradient detachment. This alignment captures where and how strongly states change without region annotations. The SFT objective is:
\begin{equation}
\mathcal{L}_{\mathrm{SFT}}
=
\mathcal{L}_{\mathrm{FM}}
+
\lambda_{\mathrm{attr}}
\mathcal{L}_{\mathrm{attr}}.
\label{eq:supervised_loss}
\end{equation}
where $\mathcal{L}_{\mathrm{FM}}$ denotes the loss of flow matching. By conditioning temporal response alignment on $\mathbf{z}_{f}$, PAN encourages the generator to associate force interventions with their induced state changes rather than rely only on correlations in the training distribution.
We further include 3,000 force free videos from WISA-80K~\citep{wang2026wisa} and drop the force embedding for $10\%$ of simulated samples to improve OOD robustness and support Stage 2. Further details are provided in Appendix~\ref{appendix_dis_pan}.

\subsection{Physics-aware Reinforcement Learning}
\label{sub:RL}

Supervised force interaction data cover only limited physical dynamics. We introduce PARL to improve physical consistency and generalization using auxiliary OOD videos. PARL combines an attribution-guided world model reward with a Dynamic Gate that filters degenerate rollouts.

Given an OOD initial state $\mathbf{I}_{\mathrm{ood}}$ and a force-free embedding $\mathbf{z}_{\varnothing}$, the old policy generates $K$ clean rollout latents $\mathbf{z}_0^{1:K}$. These latents are decoded into videos $\mathbf{X}^{1:K}$ for reward evaluation and forward noised into $\mathbf{z}_t^{1:K}$ for DiffusionNFT optimization.

\paragraph{Attribution-aligned physics reward.}
Following~\citep{yuan2026inference}, we divide each decoded video $\mathbf{X}^k$ into $C$ context frames and $T$ future frames. For each window $\ell$, the frozen V-JEPA predictor $P_{\phi}$ estimates the future features from the context, while its encoder $E_{\psi}$ extracts the corresponding target features from the complete window:
\begin{equation}
\hat{\mathbf{H}}_{\ell}^{k,\mathrm{fut}}
=
P_{\phi}\!\left(
E_{\psi}(\mathbf{X}_{\ell-C+1:\ell}^{k}),
\Delta_t
\right),
\qquad
\mathbf{H}_{\ell}^{k,\mathrm{fut}}
=
\left[
E_{\psi}(\mathbf{X}_{\ell-C+1:\ell+T}^{k})
\right]_{\Delta_t},
\label{eq:vjepa_prediction}
\end{equation}
where $\Delta_t$ specifies the future-token positions. The predicted and target features are arranged in their original spatiotemporal token layouts for subsequent attribution weighting.

Following Sec.~\ref{sub:attritution}, PAN produces a signed latent response
$\hat{\mathbf{a}}^k\in
\mathbb{R}^{B\times C_a\times T_a\times H_a\times W_a}$
from each clean rollout. We convert it into a nonnegative attribution map by taking the channelwise $L_2$ norm:
\begin{equation}
\hat{\mathbf{a}}^k
=
\operatorname{PAN}
\left(
\Delta\mathbf{z}_0^k,
\operatorname{sg}(\mathbf{z}_{\varnothing})
\right),
\qquad
A^k
=
\left(
\sum_{c=1}^{C_a}
\left|
\hat{a}^k
\right|^2
\right)^{\frac{1}{2}},
\label{eq:pan_scalar_map}
\end{equation}
Thus, $\mathbf{A}^k\geq0$ measures the response magnitude at each latent location. 
%We apply no softmax, min-max normalization, thresholding, or clipping. Instead, its scale is removed by the weighted average in Eq.~\ref{eq:physics_region_reward}.
We resize and temporally pool the PAN attribution map $\mathbf{A}^k$ to match the spatial patches and temporal tubelets of the V-JEPA features. The resulting nonnegative weight map $\mathbf{W}^k$ assigns an attribution weight to each predicted future token.
Thus, each V-JEPA token is assigned a nonnegative attribution weight.

The aligned map is broadcast over the feature dimension and applied to the V-JEPA prediction error:
\begin{equation}
\mathcal{L}_{\mathrm{attr}}^k
=
\frac{
\left\|
\mathbf{W}^k\odot
\left(
\hat{\mathbf{H}}_{\ell}^{k,\mathrm{fut}}-\mathbf{H}_{\ell}^{k,\mathrm{fut}}
\right)
\right\|_1
}{
D\left\|\mathbf{W}^k\right\|_1+\epsilon
},
\qquad
R_{\mathrm{phy}}^k
=
-\mathcal{L}_{\mathrm{attr}}^k,
\label{eq:physics_region_reward}
\end{equation}
This formulation evaluates local predictability at the V-JEPA token level, assigning higher rewards to predictable state changes in PAN-identified regions.

\paragraph{Dynamic Gate.}
The physics reward may assign high scores to degenerate rollouts, including static videos and motion caused by visual artifacts. We therefore introduce a reward-independent Dynamic Gate implemented by a VBench Judger. For each rollout $\mathbf{X}^k$, it evaluates the dynamic ratio $d^k$, temporal stability $q_{\mathrm{temp}}^k$, and visual quality $q_{\mathrm{vis}}^k$:
\begin{equation}
m^k
=
\mathbb{I}
\left[
\tau_d^{-}\leq d^k\leq\tau_d^{+}
\land q_{\mathrm{temp}}^k\geq\tau_{\mathrm{temp}}
\land q_{\mathrm{vis}}^k\geq\tau_{\mathrm{vis}}
\right],
\qquad
\mathcal{N}
=
\left\{
k\mid
m^k=1
\right\}.
\label{eq:dynamic_gate}
\end{equation}
The thresholds are calibrated on held-out SFT rollouts and remain fixed during reinforcement learning. The additional tolerance $\epsilon_A$ excludes empty or near-zero attribution maps. Only rollouts in $\mathcal{N}$ participate in reward normalization and optimization. The gate therefore prevents static collapse, excessive motion, visual artifacts, and invalid attribution maps from exploiting the physics reward.

\paragraph{Forward-process reinforcement learning.}
Following DiffusionNFT~\citep{zheng2025diffusionnft}, we convert the physics rewards within the eligible set into relative optimality probabilities:
\begin{equation}
\bar{R}_c
=
\frac{1}{|\mathcal{N}|}
\sum_{j\in\mathcal{N}}R_{\mathrm{phy}}^j,
\qquad
r(\mathbf{X}_0^k,c)
=
\frac{1}{2}
+
\frac{
R_{\mathrm{phy}}^k-\bar{R}_c
}{
2\max_{j\in\mathcal{N}}
\left|R_{\mathrm{phy}}^j-\bar{R}_c\right|
+\epsilon
},
\quad k\in\mathcal{N}.
\label{eq:optimality_reward}
\end{equation}
Here, rewards above and below the group mean are mapped above and below $1/2$, respectively. Conditions with fewer than two eligible rollouts are skipped, and rejected rollouts are excluded from both normalization and optimization.
Noise is added to each eligible latent $\mathbf{z}_0^k$ to obtain $\mathbf{z}_t^k$, which is optimized using Eq.~\ref{eq:implicit-policies} and Eq.~\ref{eq:diffusion-nft-loss}. The data-collection predictor $v_{\mathrm{old}}$ is updated from $v_\theta$ through exponential moving average.
\section{Experiment}

\subsection{Experimental Settings}
\noindent\textbf{\emph{Implementation Details.}}
PhysForce builds on the DiT-based Wan2.2-Fun-5B-Control model~\citep{peebles2023scalable,wan2025wan} and is trained on eight NVIDIA RTX PRO 6000 GPUs. Stage 1 jointly optimizes LFIE, PAN, and a rank-64 DiT LoRA for 4,600 steps at up to $832\times480$ resolution. Stage 2 applies DiffusionNFT~\citep{zheng2025diffusionnft} for 55 updates using a rank-32 policy LoRA initialized from the frozen Stage 1 model. Each update uses two conditions and six 30-step rollouts per condition. Further details are provided in Appendix~\ref{appendix_Imple}.

\renewcommand{\arraystretch}{0.88}
\setlength{\tabcolsep}{2.5pt}

\begin{table*}[t]
\centering
\caption{
Quantitative comparison and ablation study on the Force Prompting
out-of-distribution benchmark.
Best and second-best results are highlighted independently.
}
\scriptsize
\resizebox{\textwidth}{!}{%
\begin{tabular}{cl*{10}{c}}
\toprule
\multirow{2}{*}{Evaluation}
& \multirow{2}{*}{Method}
& \multicolumn{5}{c}{\textbf{Force Alignment}}
& \multicolumn{5}{c}{\textbf{Video Quality}} \\
\cmidrule(lr){3-7}
\cmidrule(lr){8-12}
&
& \makecell[c]{DCS $\uparrow$}
& \makecell[c]{DA@30$^\circ$ $\uparrow$}
& \makecell[c]{DA@45$^\circ$ $\uparrow$}
& \makecell[c]{MRR $\uparrow$}
& \makecell[c]{RMM $\uparrow$}
& \makecell[c]{CLIP-T $\uparrow$}
& \makecell[c]{PickScore $\uparrow$}
& \makecell[c]{Motion\\Smoothness $\uparrow$}
& \makecell[c]{Aesthetic\\Quality $\uparrow$}
& \makecell[c]{Imaging\\Quality $\uparrow$}
\\

\midrule
\multirow{4}{*}{\makecell[c]{Qualitative\\Evaluation}}
& Wan-Move
& 0.333
& 0.392
& 0.499
& 0.210
& 0.206
& 0.295
& 19.942
& \best{0.997}
& 47.413
& 62.237
\\
& Force Prompting
& \secondbest{0.575}
& \secondbest{0.714}
& \secondbest{0.748}
& \secondbest{0.810}
& \secondbest{1.453}
& \secondbest{0.306}
& \secondbest{20.797}
& \secondbest{0.996}
& \secondbest{47.568}
& \secondbest{62.238}
\\
& GoalForce
& 0.392
& 0.490
& 0.560
& 0.389
& 0.469
& 0.301
& 20.046
& \secondbest{0.996}
& 47.498
& 62.059
\\
& \textbf{PhysForce (Ours)}
& \best{0.782}
& \best{0.835}
& \best{0.863}
& \best{0.952}
& \best{2.488}
& \best{0.315}
& \best{21.065}
& \secondbest{0.996}
& \best{49.048}
& \best{64.352}
\\

\midrule
\multirow{5}{*}{\makecell[c]{Ablation Study\\Stage 1}}
& S1: Wan2.2-5B Base
& 0.010
& 0.190
& 0.252
& 0.608
& 1.154
& \best{0.316}
& \best{21.144}
& \best{0.997}
& \best{48.803}
& \best{62.833}
\\
& S2: S1 w/ LFIE
& \secondbest{0.614}
& 0.672
& \secondbest{0.768}
& \best{0.908}
& 1.987
& 0.308
& 20.830
& 0.995
& 47.320
& 62.420
\\
& S3: S2 w/ Direct $\Delta z$ Loss
& 0.606
& \secondbest{0.697}
& \secondbest{0.768}
& 0.880
& \secondbest{2.032}
& 0.310
& 20.849
& 0.995
& 47.432
& 62.440
\\
& S4: S1 w/ PAN
& 0.017
& 0.204
& 0.286
& 0.653
& 1.437
& \secondbest{0.314}
& \secondbest{20.970}
& \secondbest{0.996}
& \secondbest{47.601}
& \secondbest{62.715}
\\
& \textbf{S5: S1 w/ LFIE+PAN (Ours)}
& \best{0.625}
& \best{0.717}
& \best{0.787}
& \best{0.908}
& \best{2.063}
& 0.309
& 20.857
& 0.995
& 47.503
& 62.406
\\

\midrule
\multirow{4}{*}{\makecell[c]{Ablation Study\\Stage 2}}
& R1: R4 w/o. OOD Data
& \secondbest{0.622}
& 0.686
& 0.773
& 0.751
& 0.892
& \secondbest{0.310}
& \secondbest{20.945}
& \best{0.997}
& 48.203
& 61.045
\\
& R2: R4 w/o. Attribution Align
& 0.592
& 0.695
& 0.754
& 0.832
& 1.540
& \secondbest{0.310}
& 20.937
& \secondbest{0.996}
& \secondbest{48.246}
& 63.248
\\
& R3: R4 w/o. Dynamic Gate
& 0.621
& \secondbest{0.717}
& \secondbest{0.779}
& \secondbest{0.933}
& \secondbest{2.205}
& 0.308
& 20.804
& 0.995
& 47.514
& \secondbest{63.265}
\\
& \textbf{R4: PARL (Ours)}
& \best{0.782}
& \best{0.835}
& \best{0.863}
& \best{0.952}
& \best{2.488}
& \best{0.315}
& \best{21.065}
& \secondbest{0.996}
& \best{49.048}
& \best{64.352}
\\

\bottomrule
\end{tabular}%
}
\label{tab:ood}
\vspace{-0.4cm}
\end{table*}
\begin{figure*}[t]
    \centering
    \includegraphics[width=0.95\textwidth]{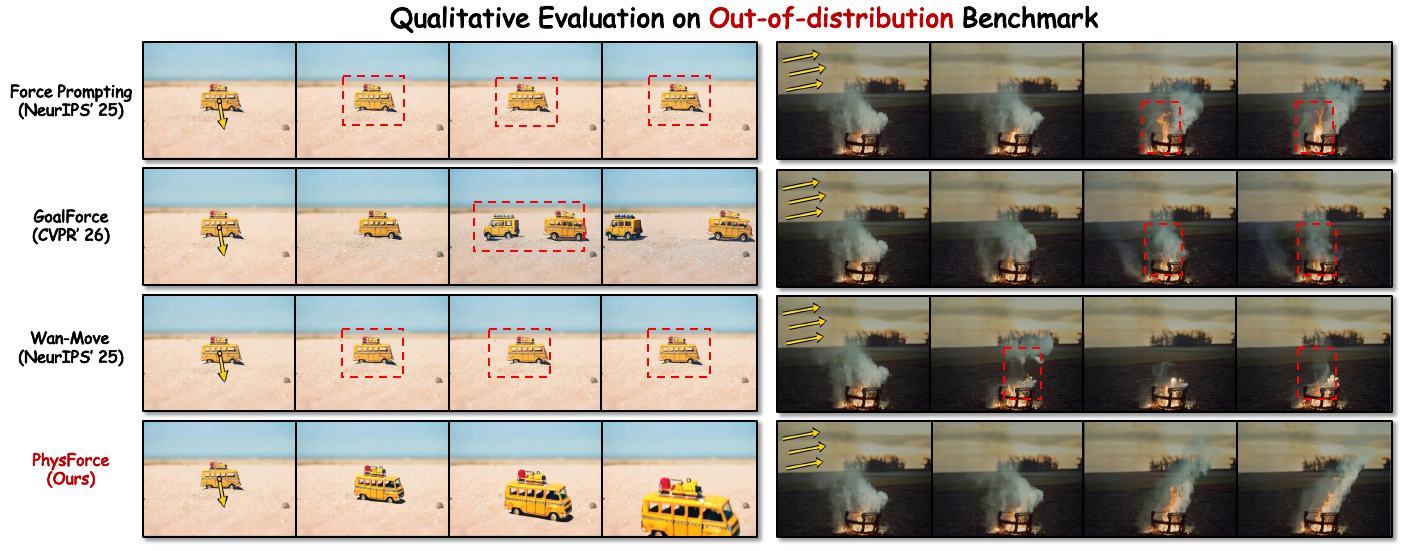}
    \vspace{-10pt}
    \caption{\textbf{Qualitative comparison of OOD Benchmark.}}
    \label{fig_comparison_ood}
    \vspace{-0.6cm}
\end{figure*}
\noindent\textbf{\emph{Dataset.}}
\textbf{Training data.}
Stage 1 uses approximately 12K point-force videos from GoalForce~\citep{gillman2026goal}, 15,359 wind-force videos from ForcePrompting~\citep{gillman2026force}, and 3K force-free videos from WISA-80K~\citep{wang2026wisa}. Stage 2 uses 100 static-camera WISA-80K videos without force annotations.
\textbf{Evaluation benchmarks.}
For OOD evaluation, we use the official ForcePrompting test set~\citep{gillman2026force}, containing 305 point-force and 52 wind-force cases disjoint from training. For ID evaluation, GT80 contains 20 held-out cases each for balls, dominoes, plants, and wind. ForcePrompting evaluates force alignment and video quality, while GT80 additionally evaluates ground-truth physics alignment. Details are provided in Appendix~\ref{appendix_dataset}.

\noindent\textbf{\emph{Metrics.}}
Following prior work~\citep{wang2026streaming,gillman2026force,gillman2026goal}, we evaluate ID and OOD performance using 14 metrics across three dimensions.
\noindent\textbf{(1) Force Alignment.}
\emph{Directional Cosine Similarity (DCS)} measures the directional agreement between the applied force and generated motion. \emph{DA@30$^\circ$} and \emph{DA@45$^\circ$} report the percentages of samples with directional errors below $30^\circ$ and $45^\circ$, respectively. \emph{Motion Response Rate (MRR)} measures the frequency of valid motion responses, while \emph{ROI Motion Magnitude (RMM)} quantifies motion strength within the affected region.
\noindent\textbf{(2) Video Quality.}
\emph{CLIP-T} and \emph{PickScore}~\citep{kirstain2023pick} evaluate text alignment and learned human preference, respectively. We further adopt \emph{Motion Smoothness}, \emph{Aesthetic Quality}, and \emph{Imaging Quality} from VBench~\citep{huang2024vbench,zheng2025vbench} to assess temporal consistency, visual appeal, and image fidelity.
\noindent\textbf{(3) Physics Alignment.}
For cases with paired ground-truth videos, we follow the Physics-IQ evaluation protocol~\citep{Motamed_2026_WACV}. \emph{Spatial IoU}, \emph{Spatiotemporal IoU (ST-IoU)}, and \emph{Weighted IoU} measure agreement between the generated and ground-truth motion regions, while \emph{MSE} measures their pixel-level discrepancy.

\renewcommand{\arraystretch}{0.90}
\setlength{\tabcolsep}{2.5pt}
\begin{table*}[t]
\centering
\caption{
Quantitative comparison and ablation study on the GT80
in-distribution benchmark.
Best and second-best results are highlighted independently.
}
\scriptsize
\resizebox{\textwidth}{!}{%
\begin{tabular}{cl*{9}{c}}
\toprule
\multirow{2}{*}{Evaluation}
& \multirow{2}{*}{Method}
& \multicolumn{5}{c}{\textbf{Force Alignment}}
& \multicolumn{4}{c}{\textbf{GT-based Physics Alignment}} \\
\cmidrule(lr){3-7}
\cmidrule(lr){8-11}
&
& \makecell[c]{DCS $\uparrow$}
& \makecell[c]{DA@30$^\circ$ $\uparrow$}
& \makecell[c]{DA@45$^\circ$ $\uparrow$}
& \makecell[c]{MRR $\uparrow$}
& \makecell[c]{RMM $\uparrow$}
& \makecell[c]{Spatial IoU $\uparrow$}
& \makecell[c]{ST-IoU $\uparrow$}
& \makecell[c]{Weighted IoU $\uparrow$}
& \makecell[c]{MSE $\downarrow$}
\\
\midrule
\multirow{4}{*}{\makecell[c]{Qualitative\\Evaluation}}
& Wan-Move
& \secondbest{0.464}
& 0.488
& 0.550
& 0.450
& 0.392
& 0.373
& 0.120
& 0.252
& 0.005
\\
& Force Prompting
& 0.433
& \secondbest{0.563}
& \secondbest{0.600}
& 0.663
& 0.856
& \secondbest{0.657}
& \secondbest{0.255}
& \secondbest{0.522}
& \secondbest{0.004}
\\
& GoalForce
& 0.433
& 0.550
& 0.575
& \secondbest{0.725}
& \secondbest{0.873}
& 0.525
& 0.186
& 0.404
& 0.006
\\
& \textbf{PhysForce (Ours)}
& \best{0.697}
& \best{0.700}
& \best{0.775}
& \best{0.763}
& \best{1.447}
& \best{0.727}
& \best{0.332}
& \best{0.614}
& \best{0.003}
\\
\midrule
\multirow{5}{*}{\makecell[c]{Ablation Study\\Stage 1}}
& S1: Wan2.2-5B Base
& 0.112
& 0.138
& 0.238
& 0.613
& 1.148
& 0.367
& 0.197
& 0.293
& 0.007
\\
& S2: S1 w/ LFIE
& \secondbest{0.509}
& \secondbest{0.575}
& \secondbest{0.663}
& \secondbest{0.755}
& \secondbest{1.206}
& 0.610
& \secondbest{0.250}
& 0.495
& \secondbest{0.005}
\\
& S3: S2 w/ Direct $\Delta z$ Loss
& \secondbest{0.509}
& 0.550
& \secondbest{0.663}
& 0.743
& 1.144
& \secondbest{0.616}
& \secondbest{0.250}
& \secondbest{0.499}
& \secondbest{0.005}
\\
& S4: S1 w/ PAN
& 0.118
& 0.263
& 0.325
& 0.663
& 0.712
& 0.367
& 0.125
& 0.260
& 0.007
\\
& \textbf{S5: S1 w/ LFIE+PAN (Ours)}
& \best{0.652}
& \best{0.638}
& \best{0.738}
& \best{0.758}
& \best{1.308}
& \best{0.654}
& \best{0.275}
& \best{0.538}
& \best{0.004}
\\
\midrule
\multirow{4}{*}{\makecell[c]{Ablation Study\\Stage 2}}
& R1: R4 w/o. OOD Data
& 0.409
& 0.550
& 0.638
& 0.438
& 0.467
& 0.488
& 0.161
& 0.365
& \secondbest{0.004}
\\
& R2: R4 w/o. Attribution Align
& 0.417
& 0.575
& 0.613
& 0.688
& 1.245
& \secondbest{0.650}
& \secondbest{0.283}
& \secondbest{0.545}
& \secondbest{0.004}
\\
& R3: R4 w/o. Dynamic Gate
& \secondbest{0.536}
& \secondbest{0.600}
& \secondbest{0.650}
& \secondbest{0.750}
& \secondbest{1.360}
& 0.634
& 0.273
& 0.519
& 0.005
\\
& \textbf{R4: PARL (Ours)}
& \best{0.697}
& \best{0.700}
& \best{0.775}
& \best{0.763}
& \best{1.447}
& \best{0.727}
& \best{0.332}
& \best{0.614}
& \best{0.003}
\\
\bottomrule
\end{tabular}%
}
\label{tab:id}
\vspace{-0.4cm}
\end{table*}
\begin{figure*}[t]
    \centering
    \includegraphics[width=0.95\textwidth]{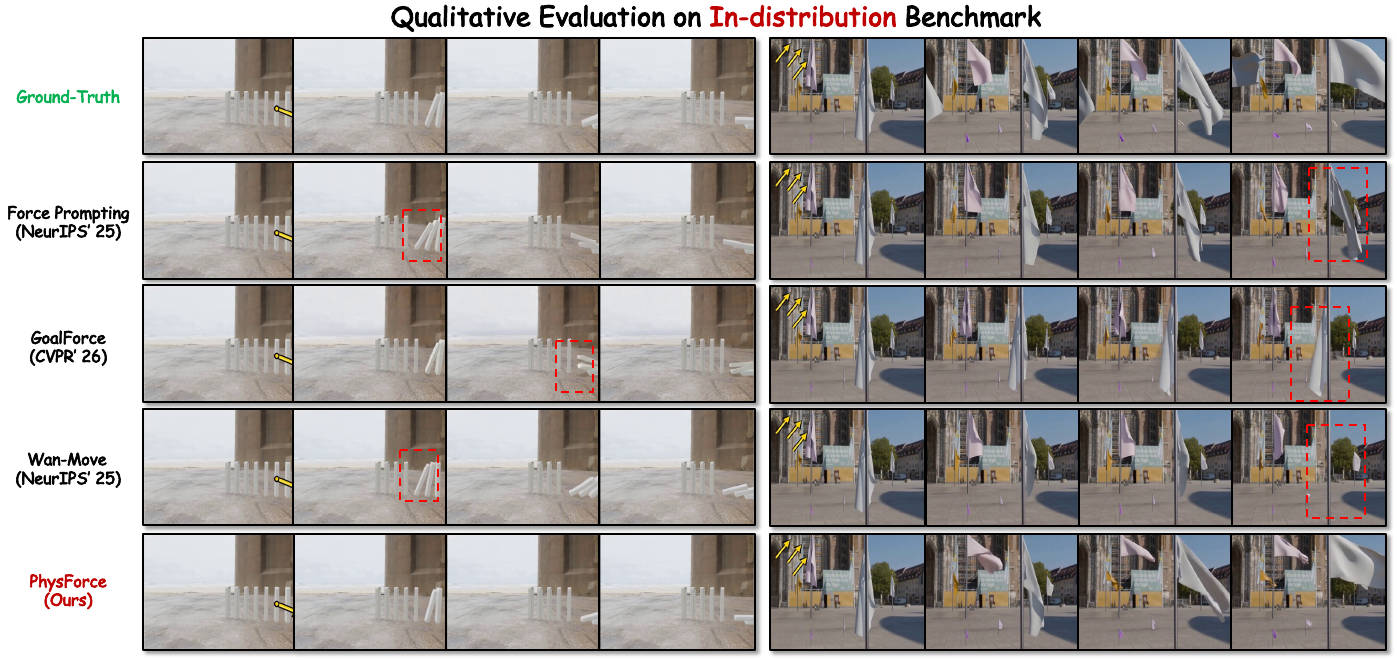}
    \vspace{-10pt}
    \caption{\textbf{Qualitative comparison of ID Benchmark.}}
    \label{fig_comparison_id}
    \vspace{-0.6cm}
\end{figure*}
\noindent\textbf{\emph{Compared Methods.}} 
To evaluate the effectiveness of PhysForce, we compare it with State-of-the-Art (SOTA) approaches. We adapt \textbf{Wan-Move}~\citep{chu2026wan}, \textbf{ForcePrompting}~\citep{gillman2026force} and GoalForce~\citep{gillman2026goal} to the Wan2.2-5B base model for fair comparison. 

\subsection{Qualitative Evaluation}
We compare PhysForce with Wan-Move~\citep{chu2026wan}, ForcePrompting~\citep{gillman2026force}, and GoalForce~\citep{gillman2026goal} in force response, video quality, and physical consistency.
\textbf{(1) OOD Comparison.}
As shown in Fig.~\ref{fig_comparison_ood}, the baselines struggle with challenging force conditions. For point force, ForcePrompting and Wan-Move produce nearly static videos, while GoalForce duplicates the vehicle. PhysForce moves the vehicle toward the foreground with coherent perspective and scale changes, capturing the implicit 3D motion. For wind force, the baselines either separate the evolution of smoke and flame or fail to model the wind response. PhysForce preserves their interaction and generates motion aligned with the wind.
\textbf{(2) ID Comparison.}
As shown in Fig.~\ref{fig_comparison_id}, the baselines also struggle with force location and magnitude. In the domino example, ForcePrompting and Wan-Move respond at incorrect locations, while GoalForce generates an inconsistent fallen domino. PhysForce reproduces the ground-truth transition. In the wind example, ForcePrompting underestimates the response, GoalForce remains nearly static, and Wan-Move produces artifacts. PhysForce captures both the waving direction and the degree of flag extension.

\subsection{Quantitative Evaluation}
We compare PhysForce with representative force-controlled video generation methods, including Wan-Move~\citep{chu2026wan}, Force Prompting~\citep{gillman2026force}, and GoalForce~\citep{gillman2026goal}. The comparison covers force alignment, video quality, and ground-truth physics alignment on both OOD and ID benchmarks.

\noindent\textbf{\emph{Objective Evaluation.}}
As shown in Tab.~\ref{tab:ood} and~\ref{tab:id}, PhysForce outperforms existing methods across the evaluated dimensions.
\textbf{(1) OOD Comparison.}
\begin{wrapfigure}{r}{0.22\textwidth}
	\centering
    \vspace{-10pt}
	\includegraphics[width=\linewidth]{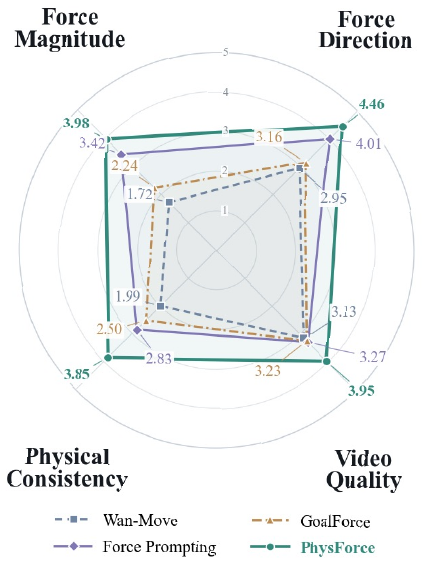}
    \vspace{-20pt}
	\caption{\footnotesize Quantitative User Study.}
    \label{fig_userstudy}
    \vspace{-0.5cm}
\end{wrapfigure}
On the OOD benchmark, PhysForce achieves the best results across all force alignment metrics. Compared with the best baseline, DCS improves from $0.575$ to $0.782$, MRR from $0.810$ to $0.952$, and RMM from $1.453$ to $2.488$. It also achieves the highest CLIP-T, PickScore, aesthetic quality, and imaging quality while maintaining comparable motion smoothness, demonstrating robust force response and video quality in unseen scenarios.
\textbf{(2) ID Comparison.}
On the GT80 benchmark, PhysForce achieves the best results across all force and physics alignment metrics. Compared with the strongest baselines, DCS improves from $0.464$ to $0.697$, RMM from $0.873$ to $1.447$, Spatial IoU from $0.657$ to $0.727$, and ST-IoU from $0.255$ to $0.332$, while MSE decreases from $0.004$ to $0.003$. These results demonstrate state changes consistent with both the applied force and ground-truth physical evolution.

\noindent\textbf{\emph{User Study.}}
We recruited 150 participants to evaluate four methods on 40 scenes, including 20 point-force and 20 wind-force cases. Participants rated force magnitude response, direction adherence, video quality, and physical consistency on a five-point scale. Method names were hidden and presentation orders randomized. As shown in Fig.~\ref{fig_userstudy}, PhysForce ranks first in all dimensions, exceeding the strongest baseline by $0.56$, $0.45$, $0.68$, and $1.02$ points, with the largest gain in physical consistency. More details in Appendix~\ref{appendix_userstudy}.

\begin{figure*}[t]
    \centering
    \includegraphics[width=1.0\textwidth]{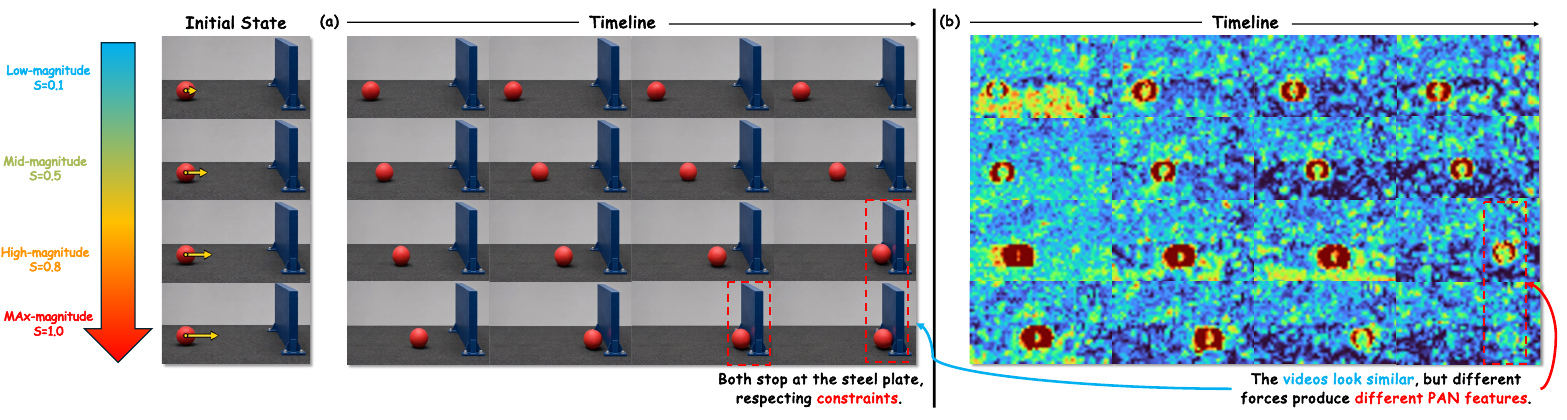}
    \vspace{-20pt}
    \caption{\textbf{Qualitative ablation study of varying force magnitudes under scene constraints.} (a) Generated video frames under different force magnitudes. (b) Corresponding PAN features, highlighting the temporal differences between ongoing motion and settled states.}
    \label{fig_ablation_constraint}
    \vspace{-0.3cm}
\end{figure*}

\begin{figure*}[!t]
    \centering
    \begin{minipage}[c]{0.55\textwidth}
        \centering
        \includegraphics[width=\linewidth]{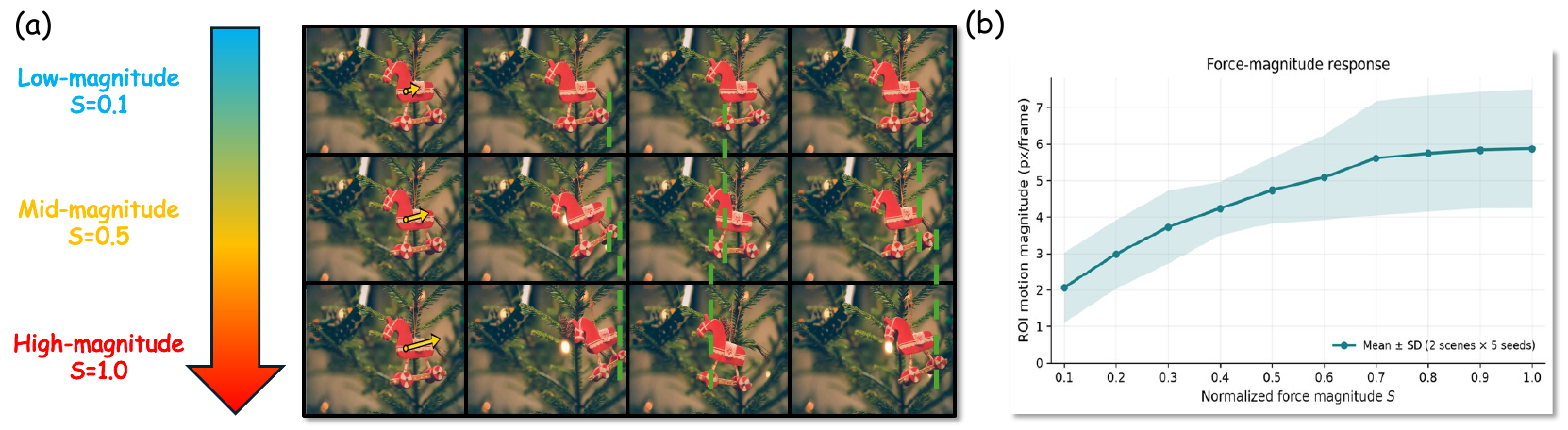}
        \captionof{figure}{\textbf{Ablation across force magnitudes.} (a) Qualitative evaluation of forces of varying magnitudes. (b) Quantitative response curve of forces of varying magnitudes.}
        \label{fig_ablation_mag}
    \end{minipage}%
    \hfill
    \begin{minipage}[c]{0.4\textwidth}
        \centering
        \begingroup
    \centering
    \scriptsize
    \captionof{table}{Comparison with ordinary motion localization methods.}
    \label{tab:pan}
    \vspace{-6pt}
    \setlength{\tabcolsep}{2pt}
    \begin{tabular*}{\linewidth}{@{\extracolsep{\fill}}lccc}
        \toprule
        Method & IoU $\uparrow$ & Dice $\uparrow$ & Centroid Error $\downarrow$ \\
        \midrule
        Optical Flow & 0.0389 & 0.0658 & 0.1154 \\
        SAM & 0.2488 & 0.3498 & 0.0587 \\
        \textbf{PAN (Ours)} & \textbf{0.3956} & \textbf{0.4713} & \textbf{0.0295} \\
        \bottomrule
    \end{tabular*}
    \endgroup

        \vspace{0.2cm}
        \includegraphics[width=\linewidth]{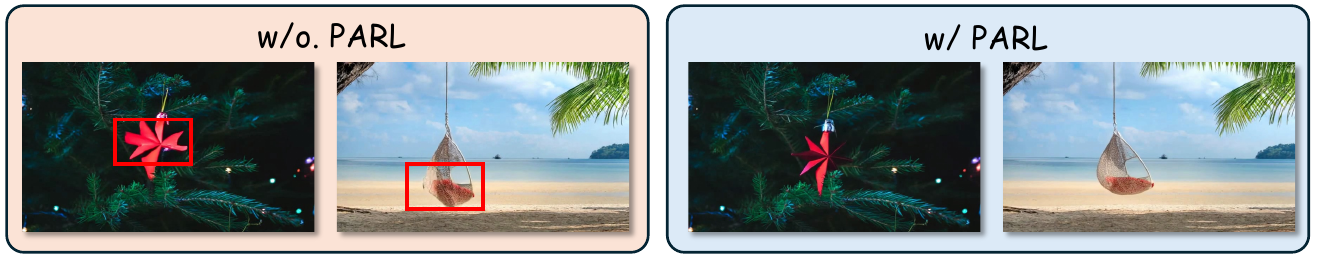}
        \captionof{figure}{Visualization of PARL results.}
        \label{fig_comparison_parl}
    \end{minipage}
    \vspace{-0.5cm}
\end{figure*}

% \begin{figure*}[t]
%     \centering
%     \begin{minipage}[t]{0.45\textwidth}
%         \centering
%         \includegraphics[width=\linewidth]{figures/extension_3Dforce.pdf}
%         \captionof{figure}{\textcolor{blue}{\textbf{Extension of our method to 3D force control.}}}
%         \label{fig_3d}
%     \end{minipage}
%     \hfill
%     \begin{minipage}[t]{0.53\textwidth}
%         \centering
%         \includegraphics[width=\linewidth]{figures/extension_temporal.pdf}
%         \captionof{figure}{\textcolor{blue}{\textbf{Extension of our method to streaming generation.}}}
%         \label{fig_streaming}
%     \end{minipage}
%     \vspace{-0.8cm}
% \end{figure*}

\subsection{Ablation Study}
We evaluate PhysForce components on the ID and OOD benchmarks in Tables~\ref{tab:id} and~\ref{tab:ood}. Additional ablations on force magnitude and direction are provided in Appendix~\ref{appendix_ablation}.

\noindent\textbf{Effect of LFIE in Stage 1. [S2]}
The base model frequently produces motion (MRR $0.608$) but shows little force alignment (DCS $0.010$). LFIE raises DCS to $0.614$, DA\@45 to $0.768$, and MRR to $0.908$, while improving ID Spatial IoU from $0.367$ to $0.610$ and Weighted IoU from $0.293$ to $0.495$. Removing VAE encoding degrades all five force-alignment metrics on both ID and OOD benchmarks (Tab.~\ref{tab:ablation_LFIE}). These gains show that LFIE transforms generic motion into force-aligned responses without prescribing future displacement. The base model retains better OOD video quality because simulation-based fine-tuning introduces a domain shift, motivating real-video refinement in Stage~2. 
We further compare LFIE with direct vector addition and cross-attention. As shown in Tab.~\ref{tab:ablation_LFIE}, LFIE consistently outperforms both alternatives across all five force-alignment metrics on both benchmarks, demonstrating the effectiveness of our conditioning design.

\noindent\textbf{Effect of PAN in Stage 1. [S3, S4, S5]}
(1) \textbf{S3} tests direct temporal-difference supervision without PAN. Despite increasing RMM, it yields marginal ID IoU gains and reduces OOD DCS from $0.614$ to $0.606$ and MRR from $0.908$ to $0.880$. Thus, direct difference matching may amplify motion without isolating force-consistent responses.
(2) \textbf{S4} evaluates PAN without LFIE. Although MRR and RMM improve, OOD DCS remains near zero. PAN models the resulting state changes but requires LFIE to provide an effective intervention signal, confirming their complementary roles.
(3) \textbf{S5} combines LFIE and PAN. With nearly unchanged MRR, ID DCS increases from $0.509$ to $0.652$ and Weighted IoU from $0.495$ to $0.538$. The gains reflect better aligned and propagated responses rather than more motion. Unlike direct difference matching, PAN learns a dedicated force-conditioned response representation that localizes how the intervention affects the scene, providing localized supervision in Stage~1 and reusable regional guidance for Stage~2 rewards.
(4) As shown in Fig.~\ref{fig_ablation_pan}, removing the force embedding substantially alters PAN's response maps. With force conditioning, PAN focuses on state changes associated with the applied intervention, demonstrating that force information guides its spatial attribution rather than merely capturing generic motion.

\noindent\textbf{Effect of OOD Data in Stage 2. [R1]}
Removing force-free OOD videos reduces OOD DCS from $0.782$ to $0.622$ and RMM from $2.488$ to $0.892$, while decreasing ID ST-IoU from $0.332$ to $0.161$ and Weighted IoU from $0.614$ to $0.365$. The degradation across both benchmarks shows that PARL benefits from exposing the shared generator to diverse real-world dynamics. The gains reflect the transfer of a refined dynamics prior rather than additional force supervision.

\noindent\textbf{Effect of Attribution Alignment in Stage 2. [R2]}
R2 tests whether global V-JEPA prediction error alone is sufficient. Without attribution alignment, static backgrounds and unrelated regions can dominate the reward. PAN-guided alignment focuses the reward on relevant state changes, improving OOD DCS from $0.592$ to $0.782$ and MRR from $0.832$ to $0.952$. It also increases ID ST-IoU from $0.283$ to $0.332$ and Weighted IoU from $0.545$ to $0.614$. PAN therefore provides the regional bridge that converts the global V-JEPA signal into a response-aware physics reward.

\noindent\textbf{Effect of the Dynamic Gate in Stage 2. [R3]}
Without the Dynamic Gate, static or degraded rollouts may exploit the predictability reward. The gate improves OOD DCS from $0.621$ to $0.782$, RMM from $2.205$ to $2.488$, and imaging quality from $63.265$ to $64.352$. It also raises ID Weighted IoU from $0.519$ to $0.614$ and reduces MSE from $0.005$ to $0.003$, demonstrating that it mitigates reward hacking while improving physical and visual consistency.

\noindent\textbf{Ablation for Force Magnitude.}
As shown in Fig.~\ref{fig_ablation_mag}, without scene constraints, increasing force magnitude produces progressively stronger motion, demonstrating continuous magnitude control. When scene constraints are introduced, as shown in Fig.~\ref{fig_ablation_constraint}(a), stronger forces drive the object to the barrier earlier, but further displacement is limited, resulting in similar final positions. Correspondingly, Fig.~\ref{fig_ablation_constraint}(b) shows that PAN captures these temporal differences, distinguishing ongoing motion from settled states.

\section{Conclusion and Future work}\label{con}
\noindent\textbf{Conclusion.}
We present PhysForce, a two-stage framework for physics-aware force control in interactive video world models. LFIE encodes forces without exposing future motion, while PAN learns localized physical responses from temporal changes. PARL further refines the dynamics prior on force-free OOD videos using PAN-guided rewards and a Dynamic Gate. Experiments show improved force alignment, physical consistency, OOD generalization, and video quality. Future work will explore explicit 3D modeling and causal streaming generation for real-time interaction.

\noindent\textbf{Future Work.}
Our future work will further explore 3D-aware force and scene representations to handle complex 3D interactions, and develop causal generation with dynamic force updates for real-time control. We will also address error accumulation and temporal drift to enable stable long-horizon generation.

\subsection*{AI Use Statement}
We used generative AI tools, including OpenAI GPT-5.6-Sol and GPT-6 Astra, to assist with methodological and experimental refinement, mathematical presentation, code development and debugging, dataset processing, translation, manuscript editing, and scientific figure preparation. 

All AI-assisted outputs were reviewed and revised by the authors. Generated data and evaluation conditions were manually inspected, AI-assisted code was tested for correctness, and all references, mathematical formulations, experimental results, and conclusions were independently verified by the authors. Generative AI was not treated as a source of empirical evidence. We take full responsibility for the final content of this work, including all text, claims, code, data, and artifacts produced with the assistance of generative AI.

\subsection*{Ethics Statement}
This work studies force-controlled video generation and physical reasoning for research purposes. While it may benefit interactive world models and robotic data generation, realistic synthetic videos could be misused to fabricate physical events or misrepresent model capabilities. We therefore recommend clearly identifying generated content and avoiding deployment in safety-critical settings without human oversight and real-world validation. Our user study involved voluntary perceptual evaluation without collecting sensitive personal information. We will follow the licenses of all source datasets and document the intended use, limitations, and potential misuse risks in any public release.

\subsection*{Reproducibility Statement}

We provide detailed descriptions of LFIE, PAN, PARL, rollout construction, training, inference, and evaluation in the main paper and appendix. We also report dataset splits, baseline adaptations, Dynamic Gate thresholds, checkpoint selection, and metric definitions. We will release the code, data-processing scripts, benchmark annotations, model configurations, and evaluation tools to support reproduction of our results.

% (This section is \textbf{recommended} and does not count toward the page limit.)

% It is important that the work published in ICLR is reproducible. Authors are
% strongly encouraged to include a paragraph-long Reproducibility Statement at the
% end of the main text (before references) to discuss the efforts that have been
% made to ensure reproducibility. This paragraph should not itself describe
% details needed for reproducing the results, but rather reference the parts of
% the main paper, appendix, and supplemental materials that will help with
% reproducibility. For example, for novel models or algorithms, a link to an
% anonymous downloadable source code can be submitted as supplementary materials;
% for theoretical results, clear explanations of any assumptions and a complete
% proof of the claims can be included in the appendix; for any datasets used in
% the experiments, a complete description of the data processing steps can be
% provided in the supplementary materials. Each of the above are examples of
% things that can be referenced in the reproducibility statement.

%\subsubsection*{Acknowledgments}

\bibliography{iclr2027_conference}
\bibliographystyle{iclr2027_conference}

\clearpage
\appendix

% Supplementary page numbering

\hypersetup{
    linkcolor=black,
    citecolor=black,
    filecolor=red,
    urlcolor=red
}

\setcounter{tocdepth}{2}
\startcontents[appendix]

\section*{Appendix}
\printcontents[appendix]{}{1}{}

\section{Discussion}\label{appendix_dis}
\subsection{Extension to Multi-Force Control}\label{appendix_dis_multi}
As illustrated in Fig.~\ref{fig:teaser}, PhysForce naturally extends to multiple simultaneous forces. In the first example, two opposing forces are applied to separate ornaments on a Christmas tree. Each ornament moves independently in the specified direction without interfering with the other. In the second example, two opposing forces of different magnitudes are applied to the ends of a domino chain. Both sides initially fall toward the center; after their effects meet, the stronger force determines the subsequent propagation direction and eventually overturns the remaining dominoes. These examples demonstrate that PhysForce can combine multiple force conditions and generate plausible interactions between their induced responses.

\begin{figure*}[!t]
    \centering
    \includegraphics[width=1.0\textwidth]{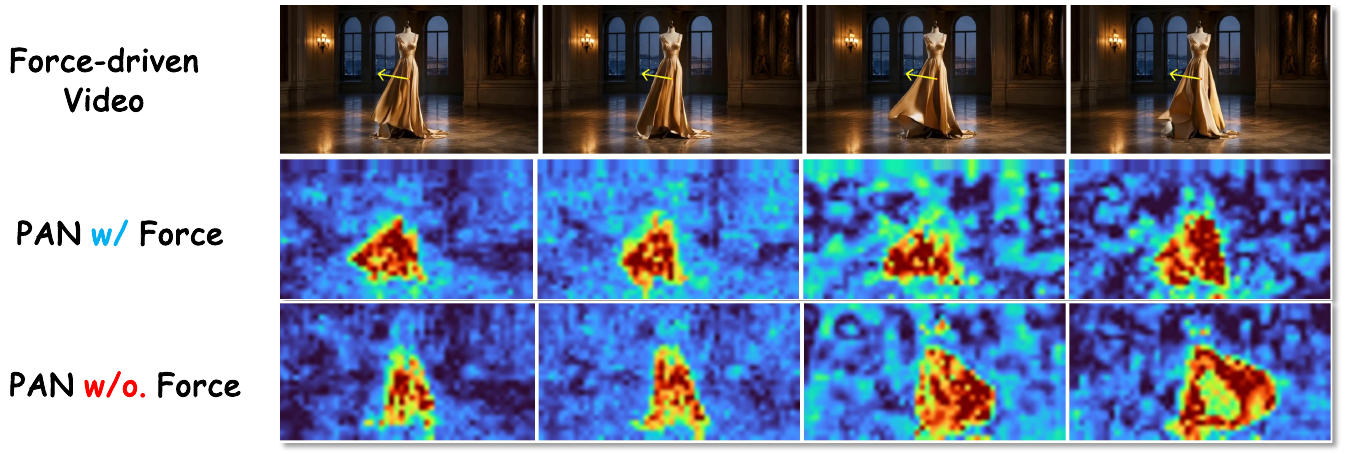}
    \caption{\textbf{Visualization of PAN} Force-diven response localization. With the force condition, PAN focuses on state changes associated with the intervention; removing it substantially alters the response map, showing that PAN uses the force information for localization.}
    \label{fig_ablation_pan}
\end{figure*}

\subsection{Why Temporal Differences Enhance Physics Attribution?}
\label{appendix_dis_pan}

PAN is primarily trained on controlled force-interaction videos. Because these videos use static cameras and the applied force is the dominant source of state change, temporal differences suppress static appearance and reveal where and how strongly the scene evolves. They provide dense supervision for the complete response, including initial motion, deformation, collisions, and subsequent motion propagation. Conditioned on $\mathbf{z}_f$, PAN transforms these differences into a localized response representation associated with the applied force.

The ablations in Tables~\ref{tab:id} and~\ref{tab:ood} show that this benefit cannot be reproduced by temporal differences alone. Compared with LFIE (S2), directly applying a $\Delta z$ loss (S3) leaves ID DCS unchanged at $0.509$ and yields only marginal gains in Spatial IoU from $0.610$ to $0.616$ and Weighted IoU from $0.495$ to $0.499$. It also reduces OOD DCS from $0.614$ to $0.606$ and MRR from $0.908$ to $0.880$. In contrast, PAN with LFIE (S5) increases ID DCS to $0.652$, Spatial IoU to $0.654$, and Weighted IoU to $0.538$, while achieving an OOD DCS of $0.625$ and DA@30 of $0.717$. Meanwhile, PAN without LFIE (S4) achieves only $0.118$ ID DCS and $0.017$ OOD DCS, confirming that PAN complements rather than replaces force conditioning.

Figure~\ref{fig_ablation_pan} further isolates the role of force information within PAN. We keep the generated video and LFIE fixed and remove only the force embedding provided to PAN. The resulting change in the response map shows that PAN uses the intervention condition rather than acting as an unconditional temporal-difference filter.

We additionally include force-free OOD videos and drop the force condition for $10\%$ of simulated samples during Stage~1. This extends the learned localization prior to scenes without explicit interventions. Conditioned on $\mathbf{z}_{\varnothing}$, PAN identifies physically relevant state changes in force-free videos, and Stage~2 uses these regions to weight the V-JEPA prediction error. The Stage~2 ablations provide further evidence: compared with removing attribution alignment (R2), PAN-guided alignment (R4) improves OOD DCS from $0.592$ to $0.782$ and MRR from $0.832$ to $0.952$. On GT80, it increases ST-IoU from $0.283$ to $0.332$ and Weighted IoU from $0.545$ to $0.614$. These results support transferring the response-localization prior learned from supervised force interactions to force-free physical refinement.

\subsection{Why V-JEPA Improves Physics Consistency?}
\label{appendix_dis_vjepa}
V-JEPA prediction error should not be interpreted as a generic measure of temporal smoothness. As a latent world model, V-JEPA predicts masked future representations from preceding observations. Learning in representation space reduces sensitivity to appearance details and encourages its features to capture object dynamics, structural continuity, interactions, and state transitions. Consequently, even visually smooth motion can receive a low reward when its future representation deviates from the dynamics learned from real videos. Prior work further shows that V-JEPA-based rewards consistently improve PhysicsIQ scores and human-perceived physical plausibility across different video generators, outperforming pixel-reconstruction and VLM-based reward signals~\citep{assran2025vjepa2,yuan2026inference}.

PhysForce transfers this learned dynamics prior through a region-aware reward. PAN restricts the V-JEPA prediction error to regions undergoing relevant state changes, preventing static backgrounds from dominating the reward, while the Dynamic Gate removes static, excessively dynamic, and visually degraded shortcuts. Our ablations further show that both attribution alignment and the Dynamic Gate improve ground-truth physics-alignment metrics. Therefore, PARL uses V-JEPA not merely to favor predictable videos, but to reinforce state evolution consistent with a learned physical dynamics prior. Although this reward is not an analytical physics solver, it provides an effective and empirically validated signal for improving physical consistency.

\subsection{How Force-Free PARL Improves Force-Conditioned Generation?}
\label{appendix_dis_PARL}
PARL does not receive explicit force conditions or optimize a reward based on force-value regression. Our premise is that Stage~1 has already learned the basic relationship between force conditions and object responses. Rather than relearning this mapping, Stage~2 uses a physics-consistency reward on force-free videos to refine the generator's dynamics prior, temporal stability, and generalization to out-of-distribution scenes.

Although Stage~2 uses the force-free embedding, both stages share the same generator parameters. Updates from PARL can therefore transfer to force-conditioned generation. Specifically, PARL evaluates the spatiotemporal consistency of object motion through V-JEPA prediction errors within PAN-localized regions, discouraging discontinuous state transitions, structural degradation, and static collapse. When force conditions are reintroduced, the model retains the condition-response mapping learned in Stage~1 while benefiting from a more stable dynamics prior for propagating the induced response.

Our qualitative results reveal two main improvements. First, once motion is initiated, the Stage~2 model is less likely to exhibit abrupt directional changes, interrupted motion, or structural collapse. Second, in unseen scenes, it better preserves motion patterns consistent with the depicted objects and is less likely to generate static videos or ignore the initial dynamic state. These observations suggest that PARL primarily improves the stability of state propagation following a force response, rather than implicitly regressing an absent external force.

The quantitative results are consistent with this interpretation. From the complete Stage~1 model to Stage~2, OOD DCS increases from $0.625$ to $0.782$, while ID ST-IoU improves from $0.275$ to $0.332$, indicating better directional force alignment in unseen scenes and closer spatiotemporal agreement with ground-truth motion regions. Meanwhile, ID MRR increases slightly from $0.758$ to $0.763$, suggesting that the improvement in ID motion-region alignment is accompanied by only a small change in motion-response frequency. Taken together, these results support a transfer benefit of physics-consistency post-training to force-conditioned generation, reflected in improved spatiotemporal motion alignment and stronger OOD generalization, rather than providing direct evidence that PARL improves force-magnitude estimation.

\section{Detailed ablation illustration}\label{appendix_ablation}
\subsection{Ablation Settings}
We detail the configurations of the ablations reported in Tables~\ref{tab:id} and~\ref{tab:ood}. Unless otherwise specified, all variants use the same backbone, training data, and optimization settings.

\begin{figure*}[!t]
    \centering
    \includegraphics[width=0.9\textwidth]{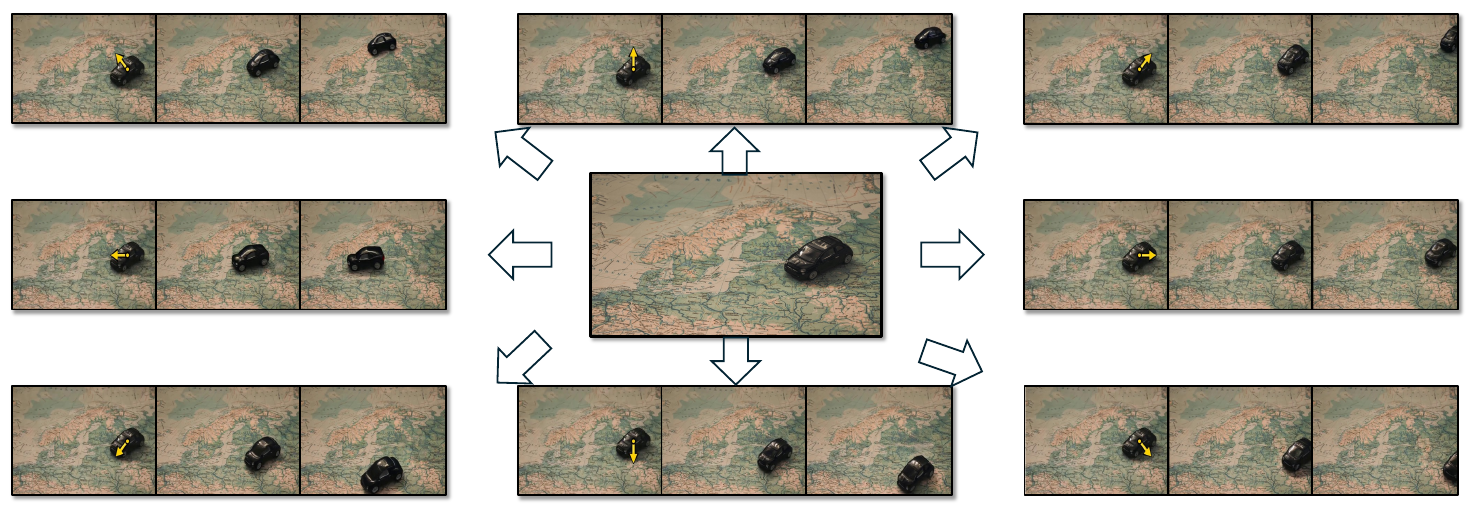}
    \caption{\textbf{Ablation across force directions.} We apply forces in eight directions to the same object and examine the resulting state changes. PhysForce consistently generates motion aligned with each applied direction, demonstrating accurate directional control.}
    \label{fig_ablation_dir}
\end{figure*}
\begin{table*}[t]
    \begingroup
    \centering
    \caption{LFIE ablation study on OOD and ID benchmarks.}
    \label{tab:ablation_LFIE}
    \scriptsize
    \resizebox{\textwidth}{!}{%
    \begin{tabular}{l*{10}{c}}
        \toprule
        \multicolumn{11}{c}{\textbf{OOD Evaluation}} \\
        \midrule
        Method & DCS $\uparrow$ & DA@30$^\circ$ $\uparrow$ & DA@45$^\circ$ $\uparrow$ & MRR $\uparrow$ & RMM $\uparrow$ & CLIP-T $\uparrow$ & PickScore $\uparrow$ & \makecell[c]{Motion\\Smoothness $\uparrow$} & \makecell[c]{Aesthetic\\Quality $\uparrow$} & \makecell[c]{Imaging\\Quality $\uparrow$} \\
        \midrule
        Vector & 0.608 & 0.705 & \secondbest{0.784} & 0.857 & 1.936 & \secondbest{0.307} & 20.805 & \secondbest{0.996} & \best{47.625} & 62.108 \\
        Cross-attn & 0.584 & 0.546 & 0.681 & 0.877 & 1.934 & 0.302 & 20.848 & \best{0.997} & 47.406 & 62.227 \\
        w/o. VAE & \secondbest{0.612} & \secondbest{0.710} & 0.723 & \secondbest{0.894} & \secondbest{1.996} & 0.305 & \secondbest{20.851} & 0.995 & \secondbest{47.517} & \secondbest{62.391} \\
        \textbf{LFIE (Ours)} & \best{0.625} & \best{0.717} & \best{0.787} & \best{0.908} & \best{2.063} & \best{0.309} & \best{20.857} & 0.995 & 47.503 & \best{62.406} \\
        \midrule\midrule
        \multicolumn{11}{c}{\textbf{ID Evaluation}} \\
        \midrule
        Method & DCS $\uparrow$ & DA@30$^\circ$ $\uparrow$ & DA@45$^\circ$ $\uparrow$ & MRR $\uparrow$ & RMM $\uparrow$ & Spatial IoU $\uparrow$ & ST-IoU $\uparrow$ & Weighted IoU $\uparrow$ & MSE $\downarrow$ & \\
        \midrule
        Vector & 0.481 & 0.537 & 0.625 & \secondbest{0.738} & 1.130 & \secondbest{0.642} & \best{0.281} & \secondbest{0.526} & \secondbest{0.005} & \\
        Cross-attn & \secondbest{0.549} & 0.537 & 0.625 & 0.726 & 1.296 & 0.575 & 0.240 & 0.460 & \secondbest{0.005} & \\
        w/o. VAE & \secondbest{0.549} & \secondbest{0.550} & \secondbest{0.633} & 0.732 & \secondbest{1.301} & 0.608 & 0.253 & 0.494 & \secondbest{0.005} & \\
        \textbf{LFIE (Ours)} & \best{0.652} & \best{0.638} & \best{0.738} & \best{0.758} & \best{1.308} & \best{0.654} & \secondbest{0.275} & \best{0.538} & \best{0.004} & \\
        \bottomrule
    \end{tabular}
    }
    \endgroup
\end{table*}

\begin{table*}[t]
    \begingroup
    \centering
    \caption{Evaluation of Force Prompting with and without reinforcement learning on OOD and ID benchmarks.}
    \label{tab:rl_forceprompting}
    \scriptsize
    \resizebox{\textwidth}{!}{%
    \begin{tabular}{l*{10}{c}}
        \toprule
        \multicolumn{11}{c}{\textbf{OOD Evaluation}} \\
        \midrule
        Method & DCS $\uparrow$ & DA@30$^\circ$ $\uparrow$ & DA@45$^\circ$ $\uparrow$ & MRR $\uparrow$ & RMM $\uparrow$ & CLIP-T $\uparrow$ & PickScore $\uparrow$ & \makecell[c]{Motion\\Smoothness $\uparrow$} & \makecell[c]{Aesthetic\\Quality $\uparrow$} & \makecell[c]{Imaging\\Quality $\uparrow$} \\
        \midrule
        Force Prompting & 0.575 & 0.714 & 0.748 & 0.810 & 1.453 & 0.306 & 20.797 & 0.996 & 47.568 & 62.238 \\
        Force Prompting w/ RL & \best{0.703} & \best{0.805} & \best{0.831} & \best{0.887} & \best{1.692} & \best{0.310} & \best{20.962} & \best{0.997} & \best{48.241} & \best{63.312} \\
        \midrule\midrule
        \multicolumn{11}{c}{\textbf{ID Evaluation}} \\
        \midrule
        Method & DCS $\uparrow$ & DA@30$^\circ$ $\uparrow$ & DA@45$^\circ$ $\uparrow$ & MRR $\uparrow$ & RMM $\uparrow$ & Spatial IoU $\uparrow$ & ST-IoU $\uparrow$ & Weighted IoU $\uparrow$ & MSE $\downarrow$ & \\
        \midrule
        Force Prompting & 0.433 & 0.563 & 0.600 & 0.663 & 0.856 & 0.657 & 0.255 & 0.522 & \best{0.004} & \\
        Force Prompting w/ RL & \best{0.529} & \best{0.627} & \best{0.668} & \best{0.701} & \best{1.092} & \best{0.724} & \best{0.276} & \best{0.586} & \best{0.004} & \\
        \bottomrule
    \end{tabular}
    }
    \endgroup
\end{table*}

\noindent\textbf{Stage~1 ablations.}
\textbf{S1} is the Wan2.2-5B base model without LFIE or PAN.
\textbf{S2} adds LFIE to evaluate the effect of force conditioning without response attribution.
\textbf{S3} removes PAN and directly applies an $\ell_1$ loss between the predicted and ground-truth latent temporal differences, testing whether direct difference matching can replace the learned response representation.
\textbf{S4} retains PAN but removes LFIE, examining response supervision without an effective force-conditioning interface.
\textbf{S5} combines LFIE and PAN as the complete Stage~1 model.

\noindent\textbf{Stage~2 ablations.}
All Stage~2 variants are initialized from S5.
\textbf{R1} removes the additional force-free OOD videos while retaining the PARL objective.
\textbf{R2} removes PAN-guided attribution alignment and globally aggregates the V-JEPA prediction error.
\textbf{R3} disables the Dynamic Gate, allowing all sampled rollouts to participate in reward normalization and optimization.
\textbf{R4} is the complete PARL model, combining force-free OOD data, PAN-guided regional rewards, and the Dynamic Gate.

\subsection{More ablation studies}
We further examine how PhysForce responds to changes in force direction and magnitude. In each experiment, we vary only one force factor while keeping the initial scene, application point, and remaining conditions fixed.

\noindent\textbf{Force Direction.}
As shown in Fig.~\ref{fig_ablation_dir}, we apply forces in eight directions to the same object with a fixed magnitude. The generated motion consistently follows the specified direction while preserving the scene content. This demonstrates that PhysForce accurately distinguishes force directions and translates them into corresponding state changes.

\noindent\textbf{Role of Force Conditioning in PAN.}
As shown in Fig.~\ref{fig_ablation_pan}(a), we keep the generated video, latent temporal differences, generator, and LFIE unchanged, and remove only the force embedding provided to PAN. With the correct force condition, PAN concentrates on localized state changes associated with the applied intervention. Replacing it with the null embedding produces a substantially different response map that captures broader object motion. Because the visual input and temporal differences are identical in both settings, this difference can only arise from PAN's force-conditioning branch. The experiment therefore shows that PAN uses the intervention information rather than functioning merely as an unconditional temporal-difference filter.

\begin{figure*}[htbp]
    \centering
    \includegraphics[width=1.0\textwidth]{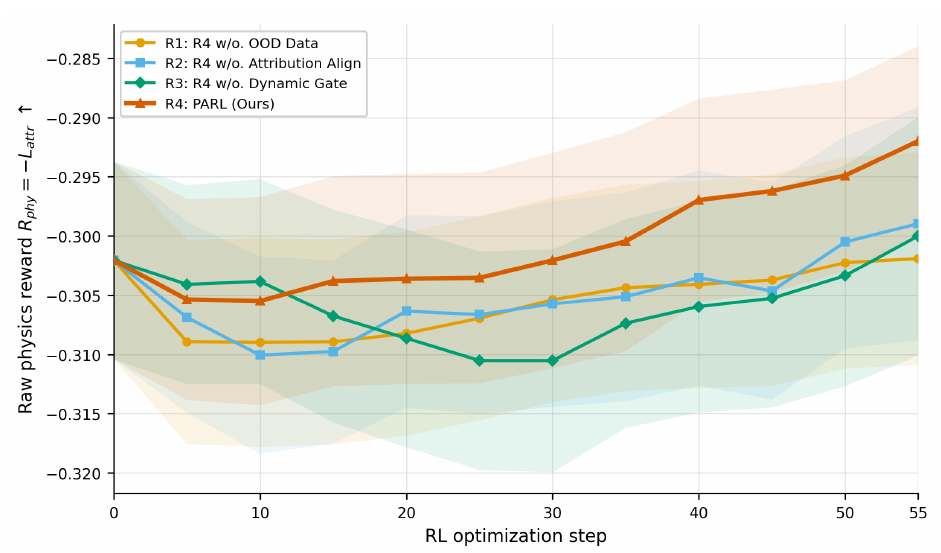}
    \caption{We visualize the reward trajectories of PARL and its ablated variants. The full model improves steadily and reaches its peak at step 55, consistently outperforming all ablations throughout optimization.}
    \label{fig_rlline}
\end{figure*}
PAN is also trained on force-free videos and simulated samples with force dropout. Consequently, when no intervention is available, it operates in a complementary force-free mode. As shown in Fig.~\ref{fig_ablation_pan}(b), PAN conditioned on $\mathbf{z}_{\varnothing}$ identifies naturally evolving regions in a completely force-free video, providing localized guidance for PARL. Thus, PAN represents force-associated responses when $\mathbf{z}_f$ is available and general physical state changes under the null condition. This evidence supports PAN as a force-conditioned response representation, although it should not be interpreted as formal counterfactual causal identification.

\noindent\textbf{Ablation of PARL Components.}
As shown in Fig.~\ref{fig_rlline}, full PARL recovers from an initial decline and steadily improves its raw physics reward from $-0.302$ to $-0.292$, reaching the highest value at update 55. Removing OOD data, attribution alignment, or the Dynamic Gate consistently yields lower final rewards. In particular, removing the Dynamic Gate causes the largest intermediate degradation and a slower recovery. During training, the Dynamic Gate retains $65.3\%$ of rollout groups on average. These results demonstrate that OOD data, attribution-guided alignment, and rollout filtering jointly improve the effectiveness and stability of PARL optimization.

\noindent\textbf{Generalization of PARL.}
As shown in Fig.~\ref{fig_ablation_dir}, we apply forces in eight directions to the same object with a fixed magnitude. The generated motion consistently follows the specified direction while preserving the scene content. This demonstrates that PhysForce accurately distinguishes force directions and translates them into corresponding state changes.

\begin{figure*}[!t]
    \centering
    \includegraphics[width=0.85\textwidth]{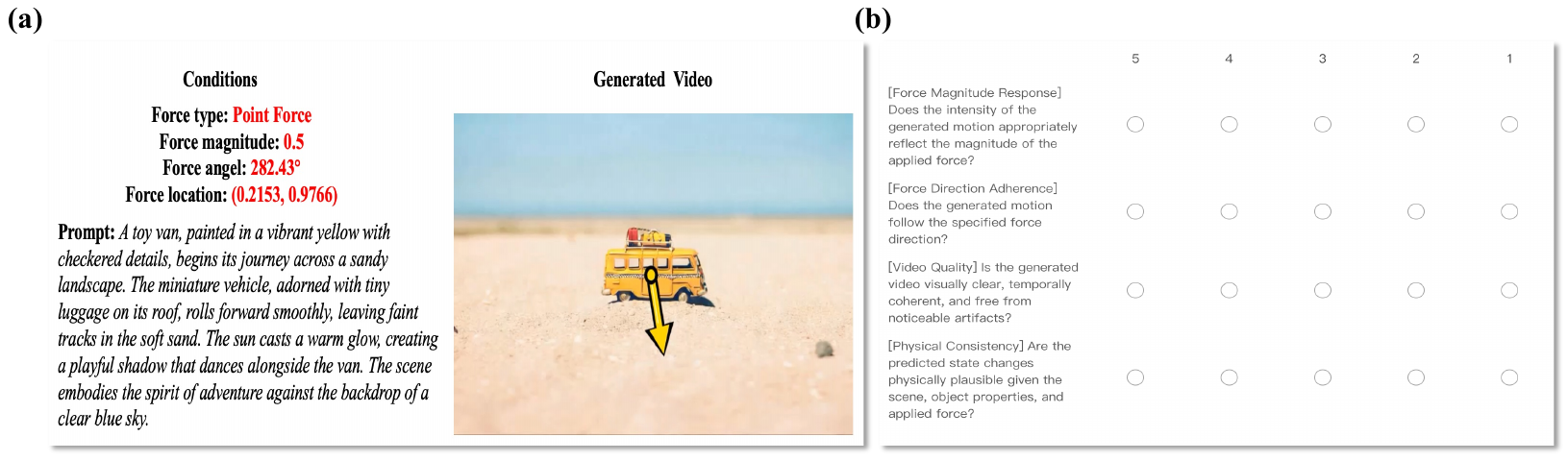}
    \caption{\textbf{Human evaluation questionnaire format.} (a) presents the scene prompt, force type, magnitude, direction, location, and generated video. (b) asks participants to rate the video from 1 to 5 in terms of force magnitude response, force direction adherence, video quality, and physical consistency.}
    \label{fig_human}
\end{figure*}

\section{Detailed User Study Instructions}\label{appendix_userstudy}
We conduct a user study to compare videos generated by Wan-Move, ForcePrompting, GoalForce, and PhysForce. Each participant evaluates 40 scenes, including 20 point force and 20 wind force scenarios. For each scene, the initial state, force condition, and four generated videos are presented together. A single arrow represents a point force, while three arrows represent wind; the arrow direction and length indicate the force direction and magnitude, respectively. Method identities are hidden and presentation orders are randomized to reduce bias. The questionnaire format is shown in Fig.~\ref{fig_human}.
Participants rate each video from 1 to 5 according to four criteria:
\textbf{(1) Force Magnitude Response.}
Whether the intensity of the generated motion appropriately reflects the magnitude of the applied force.
\textbf{(2) Force Direction Adherence.}
Whether the generated motion follows the specified force direction.
\textbf{(3) Video Quality.}
Whether the video maintains visual fidelity, temporal coherence, and freedom from noticeable artifacts.
\textbf{(4) Physical Consistency.}
Whether the predicted state changes are physically plausible given the scene, object properties, and applied force.

\section{Additional Implementation Details}\label{appendix_Imple}
\noindent\textbf{PARL Training and Checkpoint Selection.}\label{appendix_Imple_PARL}
We select WISA-80K videos using predefined criteria that require a static camera, observable object motion, and no severe visual artifacts. The ID and OOD test benchmarks are not used for video selection, Dynamic Gate calibration, hyperparameter tuning, or checkpoint selection.

PARL is trained for 100 updates, with checkpoints evaluated every five updates on the fixed validation split. We select the checkpoint at update 55 because it achieves the highest mean validation physics-consistency reward, after which the reward plateaus without further improvement. This checkpoint is then evaluated once on the ID and OOD test benchmarks. Figure~\ref{fig_rlline} reports the mean reward, with the shaded regions indicating standard deviation/95\% confidence intervals. All variants use the same data split, initialization, rollout seeds, and checkpoint-selection protocol.

\begin{figure*}[htbp]
    \centering
    \includegraphics[width=1.0\textwidth]{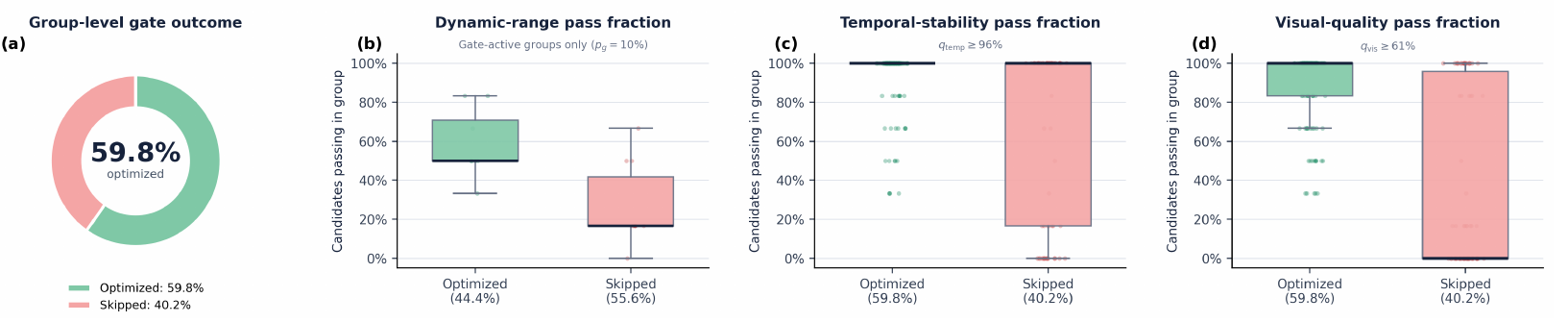}
    \caption{\textbf{Analysis of the Dynamic Gate.} Up to the selected step-55 checkpoint, the gate retains $59.8\%$ of rollout groups for optimization and skips $40.2\%$. The remaining panels compare the dynamic-range pass fraction for gate-active groups and the temporal-stability and visual-quality pass fractions for all groups, showing that the gate effectively excludes degenerate rollouts.}
    \label{fig_dynamic}
\end{figure*}
\noindent\textbf{Dynamic Gate.}\label{appendix_Imple_Dyn}
We analyze the Dynamic Gate in Fig.~\ref{fig_dynamic}. For every rollout, the VBench Judger measures temporal stability and visual quality. It additionally computes a dynamic ratio by normalizing the rollout motion intensity against that of the frozen Stage-II policy under the same condition. A rollout is eligible if its temporal stability exceeds $96\%$, its visual quality exceeds $61\%$, and, when the dynamic criterion is active, its dynamic ratio lies within $[60\%,150\%]$. A six-candidate group is optimized only if at least two rollouts are eligible; otherwise, it is skipped.

All thresholds are calibrated on held-out SFT-policy rollouts and remain fixed throughout PARL. Up to the selected step-55 checkpoint, 110 of 184 attempted groups are optimized, corresponding to a retention rate of $59.8\%$. As shown in Fig.~\ref{fig_dynamic}, skipped groups exhibit substantially lower pass fractions under the active validity criteria. The gate neither rescales the physics reward nor operates during inference; it only prevents invalid rollout groups from contributing gradients, thereby mitigating reward hacking caused by static, excessively dynamic, flickering, or visually degraded samples.

\noindent\textbf{\emph{Compared Methods.}}\label{appendix_Imple_com}
We compare PhysForce with Wan-Move~\citep{chu2026wan}, ForcePrompting~\citep{gillman2026force}, and GoalForce~\citep{gillman2026goal}. For a controlled comparison, all methods are implemented on the same Wan2.2-5B backbone and trained using identical data splits. Each method receives the same initial image, text prompt, and force annotation, including the application point, direction, and magnitude. The force annotation is converted into the conditioning representation required by each method following its original design.

We preserve the conditioning modules and objectives of each baseline while keeping the remaining settings identical. All methods are trained at $832\times480$ resolution for 4,600 steps using the same sampling strategy, a learning rate of $1\times10^{-4}$, and a force-condition dropout probability of $10\%$. LFIE, PAN, and PARL are used only in PhysForce.

During inference, all methods generate videos at $832\times480$ resolution using 50 denoising steps and a classifier-free guidance scale of $5.0$.

\noindent\textbf{Stage 1: Force-conditioned supervised fine-tuning.}\label{appendix_Imple_stg1}
In Stage 1, we introduce the Latent Force Intervention Encoder (LFIE) and Physics Attribution Network (PAN). LFIE derives the intervention magnitude from an acceleration-related proxy. According to Newton's second law, $a=F/m$, and the initial velocity change over a fixed interval is proportional to acceleration. We therefore define the raw intervention strength as
\begin{equation}
\tilde{s}=
\begin{cases}
F/m, & \text{for balls with mass annotations},\\
F, & \text{for dominoes and plants},\\
F_{\mathrm{wind}}, & \text{for wind}.
\end{cases}
\label{eq:raw_magnitude}
\end{equation}
For balls, $F/m$ directly represents the applied acceleration. For dominoes and plants, object mass is not exposed by the simulator, so we use its native force parameter, with the resulting response jointly determined by the simulated physical properties. Similarly, $F_{\mathrm{wind}}$ denotes the native wind-control strength.

Because these quantities have different physical meanings and units, we normalize them independently within each dataset subset $\mathcal{D}$:
\begin{equation}
s=
\operatorname{clip}\!\left(
\frac{\tilde{s}-\tilde{s}_{\min}^{\mathcal{D}}}
{\tilde{s}_{\max}^{\mathcal{D}}-\tilde{s}_{\min}^{\mathcal{D}}+\epsilon},
0,1
\right).
\label{eq:magnitude_normalization}
\end{equation}
The normalization statistics are computed only from the corresponding training split. Therefore, $s$ is a dimensionless intervention strength that preserves relative magnitudes within each physical domain, rather than a shared force unit across different categories.

After converting the source direction into image coordinates, we denote its unit vector as
$\mathbf{d}=(d_x,d_y)=(\cos\theta,\sin\theta)$. Given an application point $\mathbf{p}$, LFIE constructs the factorized force field
\begin{equation}
\mathcal{F}(\mathbf{p})
=
\left[
M(\mathbf{p}),
W(\mathbf{p})d_x,
W(\mathbf{p})d_y,
W(\mathbf{p})s
\right],
\label{eq:factorized_force}
\end{equation}
where $M$ is a truncated Gaussian centered at $\mathbf{p}$ and $W$ is its binary support. Their spatial extents are fixed and independent of $s$. For global wind, we set $M=W=1$.

The force field is broadcast across the temporal latent dimension without changing its spatial support. This broadcasting describes how the condition is provided to the generator rather than how long the physical force is applied. A localized point force represents an impulse at the beginning of the interaction, whereas global wind remains active throughout the video. The current representation does not explicitly control arbitrary onset times or force durations. Since its center remains fixed, the condition contains no future displacement or object trajectory.

LFIE projects $\mathcal{F}$ into the video latent space through two complementary branches. The frozen branch constructs the magnitude-aware Cartesian components
\begin{equation}
F_x=Msd_x,\qquad F_y=Msd_y,
\end{equation}
which are converted into the native control format and encoded by the frozen video VAE to obtain the Directional Force Embedding $\mathbf{z}_{\mathrm{DFE}}$. In parallel, a trainable adapter $\mathcal{A}_{\psi}$ with a hidden dimension of 32 projects the complete four-channel field onto the 48-channel Wan2.2 latent grid. The final force embedding is
\begin{equation}
\mathbf{z}_f
=
\mathbf{z}_{\mathrm{DFE}}
+
\mathcal{A}_{\psi}(\mathcal{F}),
\label{eq:force_embedding}
\end{equation}
where the output projection of $\mathcal{A}_{\psi}$ is initialized to zero.

PAN has a hidden dimension of 64 and aligns the predicted inter-frame latent response with the corresponding ground-truth latent difference using an $\ell_1$ objective. We use a temporal interval of one and set the attribution-loss weight to $\lambda_{\mathrm{attr}}=0.02$. The LFIE adapter, PAN, and rank-64 DiT LoRA modules are jointly optimized, while the remaining DiT parameters are frozen. LoRA modules are inserted into the query, key, value, and output projections of every self-attention layer and the two feed-forward projections.

This phase is trained for 4,600 optimization steps using AdamW with a learning rate of $1\times10^{-4}$, a weight decay of $0.01$, a cosine schedule, 100 warm-up steps, and four-step gradient accumulation. We sample 49-frame and 81-frame clips with probabilities of $0.7$ and $0.3$, respectively, at a fixed resolution of $832\times480$. Clips always start from the first frame, and the force condition  are independently dropped with probability $10 \%$. Force-free WISA-80K samples use an all-zero force field.

\noindent\textbf{Stage 2: Physics-Aware Reinforcement Learning.}\label{appendix_Imple_stg2}
PARL is initialized from the 4,600-step Stage-1 checkpoint and optimized with DiffusionNFT. We use 100 camera-stationary, force-free videos selected from WISA-80K. Each rollout contains 49 frames at $832\times480$ resolution and is generated using 30 denoising steps, a classifier-free guidance scale of $5.0$, and a flow-shift value of $5.0$. Each condition produces six candidate rollouts, and two condition groups are processed per update, yielding 12 rollouts per optimization step.

We train an additional rank-32 policy LoRA with an alpha of 64 and zero dropout. AdamW is used with a constant learning rate of $5\times10^{-5}$, a weight decay of $1\times10^{-4}$, $(\beta_1,\beta_2)=(0.9,0.999)$, and a maximum gradient norm of $1.0$. The DiffusionNFT interpolation coefficient is $\beta=0.1$, and the reference regularization coefficient is $1\times10^{-4}$. Rewards are normalized within each condition group.

The physics reward is computed using a frozen V-JEPA~2.1 model. Each temporal window contains 16 frames, comprising eight context frames and eight prediction frames. A stride of eight produces five windows for each 49-frame rollout. Before encoding, the longest spatial side is resized to 384 pixels, followed by the native V-JEPA~2.1 preprocessing. The reward is defined as the negative $\ell_1$ prediction error between the predictor and target-encoder features.

A VBench-based Dynamic Gate prevents degenerate rollouts from exploiting this predictability reward. Temporal stability and imaging quality are evaluated for every rollout, with thresholds of $0.96$ and $0.61$, respectively. When activated, an eligible rollout must have a dynamic ratio between $0.6$ and $1.5$ relative to the frozen Stage-1 policy under the same condition. Groups with fewer than two eligible rollouts are skipped.

Within each valid group, physics rewards are centered over the eligible rollouts and continuously mapped to optimality probabilities following Eq.~\ref{eq:optimality_reward}. DiffusionNFT optimization is performed only over these eligible rollouts, while rejected rollouts contribute neither to the group statistics nor to the training objective.

Separately, we monitor drift from the DiffusionNFT regularization reference and refresh it when the drift exceeds $0.03$, subject to a minimum interval of three updates and a maximum interval of ten updates. PARL is trained for 100 updates, with checkpoints evaluated every five updates on a fixed validation split. The update-55 checkpoint achieves the highest validation physics reward and is used once for the reported ID and OOD evaluations.

\section{Detailed Metric Definitions} \label{appendix_metric}
We evaluate the generated videos using 14 metrics grouped into force alignment, video quality, and physics alignment.

\noindent\textbf{(1) Force Alignment.}
These metrics evaluate whether the generated motion exhibits a valid response to the input force and whether its initial direction agrees with the specified force direction. For a local point force, motion is estimated within the force-affected region using dense optical flow, after subtracting the median background flow to compensate for global camera translation. For a global force, motion is evaluated over the entire frame. Directional metrics are computed over the initial temporal window to avoid penalizing physically plausible direction changes caused by subsequent collisions or interactions.

\textbf{Directional Cosine Similarity (DCS)} is the cosine similarity between the dominant initial motion vector and the target force vector. A larger value indicates stronger directional agreement. \textbf{Direction Accuracy at $30^\circ$ (DA@30)} and \textbf{$45^\circ$ (DA@45)} report the proportions of videos whose angular errors are no greater than $30^\circ$ and $45^\circ$, respectively. \textbf{Motion Response Rate (MRR)} measures the proportion of videos whose average motion magnitude within the evaluated region exceeds a predefined response threshold. \textbf{ROI Motion Magnitude (RMM)} reports the average optical-flow displacement within the force-affected region in pixels per frame, reflecting the strength of the generated motion response.

\noindent\textbf{(2) Video Quality.}
We evaluate both semantic consistency and perceptual video quality. \textbf{CLIP-Text Score (CLIP-T)} is the mean cosine similarity between the text prompt and the individual generated frames. \textbf{PickScore} measures text-conditioned human-preference alignment by averaging the frame-level preference scores produced by the PickScore model. \textbf{Motion Smoothness} follows the AMT-based VBench implementation, which predicts intermediate frames between neighboring frames and compares them with the observed frames to evaluate temporal interpolation consistency. \textbf{Aesthetic Quality} measures the visual appeal of the generated content using the VBench aesthetic predictor. \textbf{Imaging Quality} evaluates low-level visual fidelity—including sharpness, distortion, and image degradation—using the VBench implementation.

\noindent\textbf{(3) Physics Alignment.}
For GT80 cases with paired ground-truth videos, we follow the metric definitions and official implementation of Physics-IQ~\citep{Motamed_2026_WACV}. Generated and reference videos are temporally aligned and resized to the same spatial resolution before evaluation. Motion masks are extracted using the official background-subtraction pipeline without optical flow or additional spatial registration.

\textit{Spatial IoU} measures the overlap between temporally aggregated motion regions. \textit{Spatiotemporal IoU (ST-IoU)} averages the frame-wise overlap between motion masks. \textit{Weighted IoU} compares motion-frequency maps, assigning greater importance to persistently active regions. \textit{Mean Squared Error (MSE)} measures the RGB difference between aligned generated and reference frames. We report these metrics separately because GT80 provides only one reference outcome per case and therefore does not support the physical-variance normalization required by the official Physics-IQ Score.

\section{Detailed Dataset Description}
\label{appendix_dataset}

\noindent\textbf{Stage 1 Training Data.}
The force-conditioned training set contains four datasets. GoalForce-Balls~\citep{gillman2026goal} contains approximately 6K videos of ball interactions, while GoalForce-Dominos and GoalForce-Plants contain approximately 3K videos each. These datasets provide localized point-force conditions. We additionally use 15,359 videos from the ForcePrompting training data~\citep{gillman2026force}, which provide global wind-force conditions. Stage 1 therefore contains approximately 27K force-conditioned videos covering both localized and global forces. Each sample provides a source video, text description, reference image, and temporally aligned force condition.

We further include 3K force-free videos from WISA-80K~\citep{wang2026wisa} to improve robustness to natural scenes and support the subsequent reward-based training. These videos do not provide force annotations or paired physical outcomes.

\noindent\textbf{Stage 2 Training Data.}
For reward-based post-training, we select static-camera videos from WISA-80K~\citep{wang2026wisa}. Videos with noticeable camera motion are excluded because camera displacement may be incorrectly interpreted as object dynamics by the reward model. Stage 2 uses neither force annotations nor paired target videos. No sample from either evaluation benchmark is used in Stage 1 or Stage 2.

\noindent\textbf{OOD Evaluation.}
Following the official ForcePrompting protocol~\citep{gillman2026force}, we use its complete OOD test set, which contains 357 cases comprising 305 point-force cases and 52 wind-force cases. This test set is disjoint from the force-conditioned training data and contains unseen scenes and objects. Each case specifies the force location, direction, and magnitude. We use this benchmark to evaluate force alignment and video quality.

\noindent\textbf{ID Evaluation.}
We construct GT80 from 80 videos held out from the four force-conditioned training domains before training. It contains 20 ball, 20 domino, 20 plant, and 20 wind cases, none of which are included in the training set. Because these cases follow the same scene and interaction distributions as the training data, GT80 serves as the in-distribution benchmark. Its paired ground-truth videos enable the evaluation of force alignment, video quality, and GT-based physics alignment.

\noindent\textbf{Demo Dataset.}
To examine generalization under more challenging conditions, we use GPT-6 Astra~\citep{singh2025openai} to construct a demo set following the ForcePrompting test format. This set is used exclusively for qualitative demonstration and is excluded from all quantitative evaluations. Examples on the project page show that PhysForce generalizes across diverse scenes and predicts plausible state changes according to object properties and applied forces.

\begin{figure*}[htbp]
    \centering
    \includegraphics[width=1.0\textwidth]{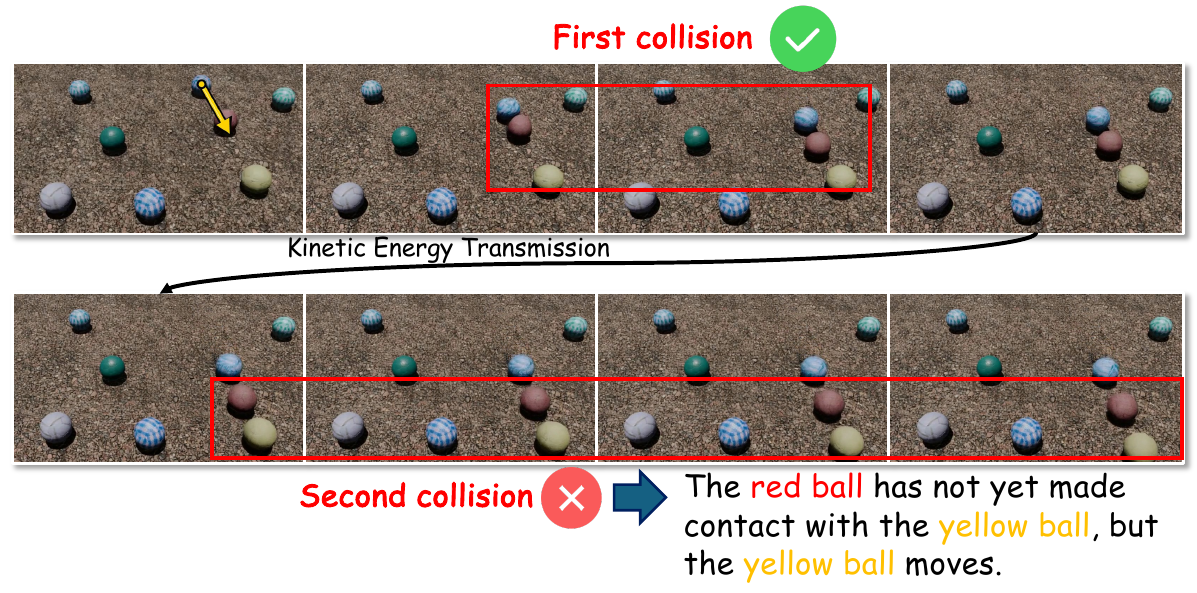}
    \caption{\textbf{Bad case of collision-related physical inconsistencies.} The first collision (top) is correctly captured, with kinetic energy transferred through physical contact. In contrast, the second collision (bottom) exhibits a nonphysical response: the yellow ball starts moving before making contact with the red ball, indicating premature energy transmission and inconsistent collision dynamics.}
    \label{fig_badcase}
\end{figure*}

\section{Limitations}\label{appendix_limitation}
\noindent\textbf{2D representation.}
PhysForce operates in the 2D video latent space. Although its outputs exhibit perspective- and scale-consistent motion, the model does not explicitly represent depth, 3D geometry, camera pose, or out-of-plane forces. It may therefore struggle with depth-dependent contact, occlusion, and complex 3D trajectories.

\noindent\textbf{Offline generation.}
PhysForce uses offline bidirectional denoising, so all force conditions must be specified before generation and cannot be updated as the scene evolves.

\noindent\textbf{Approximate physical consistency.}
PhysForce does not explicitly enforce momentum conservation or contact dynamics. V-JEPA-guided RL improves temporal consistency, but complex multi-body interactions may still exhibit artifacts, such as motion occurring before contact because of premature energy transmission, such as Fig.~\ref{fig_badcase}. Limited coverage of such interactions in existing datasets remains a key challenge.

\end{document}